%% file: iclr2024_conference.tex
\documentclass{article} 
\usepackage{iclr2024_conference,times}

\input{math_commands.tex}

\usepackage{array}
\usepackage{algorithm}
\usepackage{algorithmic}
\usepackage{amsthm,amsmath,amssymb}
\usepackage{bbm}
\usepackage{makecell}
\usepackage{booktabs}
\usepackage{color} 
\usepackage{tikz}
\usepackage{hyperref}
\hypersetup{
	colorlinks=true,
        breaklinks=true,
        citecolor={green!60!blue},
}
\usepackage{tabularx}
\usepackage{graphbox}
\usepackage{enumitem}
\usepackage{multirow}
\usepackage{mathrsfs}
\usepackage{url}
\usepackage{wrapfig}
\usepackage{subcaption}
\usepackage{adjustbox}
\usepackage[labelsep=period,labelfont=bf]{caption}

\usepackage[accsupp]{axessibility}
\newcommand{\expbf}[1]{+#1$\%$}

\newcommand{\mcbed}{\includegraphics[scale=0.4,align=c]{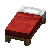}}
\newcommand{\mcbeef}{\includegraphics[scale=0.4,align=c]{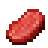}}
\newcommand{\mcbowl}{\includegraphics[scale=0.4,align=c]{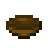}}
\newcommand{\mcbucket}{\includegraphics[scale=0.4,align=c]{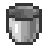}}
\newcommand{\mccarpet}{\includegraphics[scale=0.4,align=c]{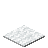}}
\newcommand{\mcchest}{\includegraphics[scale=0.4,align=c]{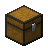}}

\newcommand{\mccobblestonewall}{\includegraphics[scale=0.4,align=c]{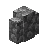}}
\newcommand{\mccookedbeef}{\includegraphics[scale=0.4,align=c]{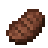}}
\newcommand{\mccookedmutton}{\includegraphics[scale=0.4,align=c]{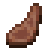}}

\newcommand{\mccraftingtable}{\includegraphics[scale=0.4,align=c]{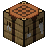}}
\newcommand{\mcdiamondsword}{\includegraphics[scale=0.4,align=c]{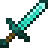}}
\newcommand{\mcfurnace}{\includegraphics[scale=0.4,align=c]{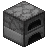}}
\newcommand{\mcironingot}{\includegraphics[scale=0.4,align=c]{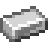}}
\newcommand{\mcitemframe}{\includegraphics[scale=0.4,align=c]{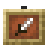}}
\newcommand{\mclever}{\includegraphics[scale=0.4,align=c]{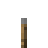}}
\newcommand{\mclog}{\includegraphics[scale=0.4,align=c]{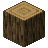}}
\newcommand{\mcmilkbucket}{\includegraphics[scale=0.4,align=c]{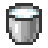}}
\newcommand{\mcmutton}{\includegraphics[scale=0.4,align=c]{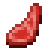}}
\newcommand{\mcpainting}{\includegraphics[scale=0.4,align=c]{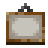}}

\newcommand{\mcshears}{\includegraphics[scale=0.4,align=c]{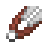}}

\newcommand{\mcsign}{\includegraphics[scale=0.4,align=c]{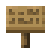}}
\newcommand{\mcstick}{\includegraphics[scale=0.4,align=c]{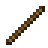}}
\newcommand{\mcstonepickaxe}{\includegraphics[scale=0.4,align=c]{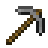}}
\newcommand{\mcstonestairs}{\includegraphics[scale=0.4,align=c]{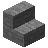}}
\newcommand{\mcstoneslab}{\includegraphics[scale=0.4,align=c]{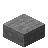}}
\newcommand{\mctorch}{\includegraphics[scale=0.4,align=c]{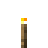}}
\newcommand{\mctrapdoor}{\includegraphics[scale=0.4,align=c]{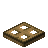}}
\newcommand{\mcwoodenpickaxe}{\includegraphics[scale=0.4,align=c]{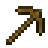}}
\newcommand{\mcwool}{\includegraphics[scale=0.4,align=c]{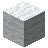}}

\newcommand{\mcheavypressureplate}{\includegraphics[scale=0.3,align=c]{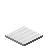}}
\newcommand{\mcironaxe}{\includegraphics[scale=0.3,align=c]{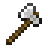}}
\newcommand{\mcironpickaxe}{\includegraphics[scale=0.3,align=c]{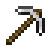}}
\newcommand{\mcironshovel}{\includegraphics[scale=0.3,align=c]{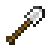}}
\newcommand{\mcironsword}{\includegraphics[scale=0.3,align=c]{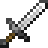}}
\newcommand{\mcirontrapdoor}{\includegraphics[scale=0.3,align=c]{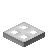}}

\newcommand{\mctripwirehook}{\includegraphics[scale=0.3,align=c]{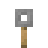}}

\newcommand{\mcdirt}{\includegraphics[scale=0.3,align=c]{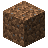}}

\makeatletter
\def\@fnsymbol#1{\ensuremath{\ifcase#1\or \dagger\or \ddagger\or
   \mathsection\or \mathparagraph\or \|\or **\or \dagger\dagger
   \or \ddagger\ddagger \else\@ctrerr\fi}}
\makeatother

\title{\includegraphics[height=15pt]{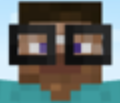} Steve-Eye: Equipping LLM-based Embodied Agents with Visual Perception in Open Worlds}

\author{Sipeng Zheng$^1$, Jiazheng Liu$^2$, Yicheng Feng$^2$, Zongqing Lu$^{1,2}$\thanks{Corresponding author}\\
$^1$ Beijing Academy of Artificial Intelligence \\
$^2$ School of Computer Science, Peking University\\
{\small \texttt{spzheng@baai.ac.cn} \quad \texttt{fyc813@pku.edu.cn} \quad\texttt{zongqing.lu@pku.edu.cn}}
}

\iclrfinalcopy 
\iclrpreprint
\begin{document}

\maketitle

\input{sections/0_abstract}
\input{sections/1_intro}

\input{sections/2_relatedwork}

\input{sections/3_method}

\input{sections/4_experiments}
\input{sections/5_conclusion}

\bibliography{iclr2024_conference}
\bibliographystyle{iclr2024_conference}

\newpage
\appendix
\input{sections/6_appendix}

\end{document}

%% file: math_commands.tex
\usepackage{amsmath,amsfonts,bm}

\def\eqref#1{equation~\ref{#1}}

\def\1{\bm{1}}

\DeclareMathAlphabet{\mathsfit}{\encodingdefault}{\sfdefault}{m}{sl}
\SetMathAlphabet{\mathsfit}{bold}{\encodingdefault}{\sfdefault}{bx}{n}

\DeclareMathOperator*{\argmax}{arg\,max}

%% file: sections/0_abstract.tex
\begin{abstract}
Recent studies have presented compelling evidence that large language models (LLMs) can equip embodied agents with the self-driven capability to interact with the world, which marks an initial step toward versatile robotics.
However, these efforts tend to overlook the visual richness of open worlds, rendering the entire interactive process akin to ``a blindfolded text-based game.''
Consequently, LLM-based agents frequently encounter challenges in intuitively comprehending their surroundings and producing responses that are easy to understand.
In this paper, we propose Steve-Eye, an end-to-end trained large multimodal model to address this limitation.
Steve-Eye integrates the LLM with a visual encoder to process visual-text inputs and generate multimodal feedback.
We adopt a semi-automatic strategy to collect an extensive dataset comprising 850K open-world instruction pairs, enabling our model to encompass three essential functions for an agent: multimodal perception, foundational knowledge base, and skill prediction and planning.
Lastly, we develop three open-world evaluation benchmarks, then carry out experiments from a wide range of perspectives to validate our model's capability to strategically act and plan.
The project’s website and code can be found at \href{https://sites.google.com/view/steve-eye}{https://sites.google.com/view/steve-eye}.
\end{abstract}


%% file: sections/1_intro.tex
\section{Introduction}

\begin{wrapfigure}{r}{0.51\textwidth} 
\vspace{-0.3cm}
\centering
  \includegraphics[width=0.50\textwidth]{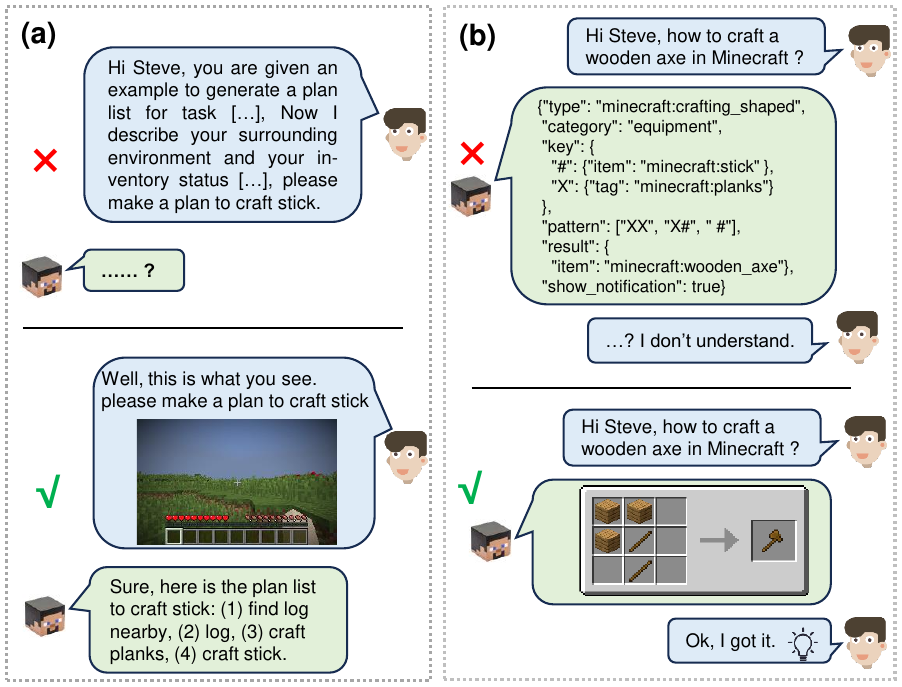}
  \vspace{-2mm}
  \caption{(a) LLM-based agent's feedback is uncontrollable due to the uncertainty of input textual prompt, while visual cues can benefit the agent to generate feedbacks; (b) a text-only driven agent often finds it difficult to produce intuitive feedback that humans can easily understand.
}
\label{fig:intro} 
\vspace{-0.3cm}
\end{wrapfigure}

Developing embodied agents that can adapt to the open world has long been a substantial challenge~\citep{kolve2017ai2,savva2019habitat}.
Recently, the rapid progress of large language models (LLMs)~\citep{chatgpt,touvron2023llama} has shown their potential to serve as a general-purpose assistant. 
Driven by these pre-trained LLMs, recently proposed agents~\citep{yuan2023plan4mc,wang2023voyager,wang2023describe,zhu2023ghost} have managed to extract world knowledge and reasoning capabilities from LLMs, allowing them to become self-driven.
Thereby these agents are capable of generating executable policies or plans for a wide range of skills and tasks in an open-ended world.

While current attempts to integrate LLMs show promise in developing a generic embodied agent, these efforts primarily translate the entire world into text, which overlooks the multifaceted richness of diverse visual reality and turns interacting with the environment into something akin to ``\textbf{a blindfolded text-based game}.''
Consequently, such text-only agents often face difficulties when it comes to effectively and intuitively representing the world. 
Imagine a situation where you request your agent to shop for a pair of shoes. 
Would you prefer to send the agent a picture of the shoes or provide a lengthy description of the shoes to convey their appearance? Undoubtedly, you would opt for the former choice.

In fact, the agent's reliance on text input/output (I/O) imposes significant limitations on its ability to interact with the world. 
To illustrate this point, we consider Minecraft~\citep{guss2019minerl,fan2022minedojo} as an ideal example.
Minecraft, being an expansive sandbox game, offers a vast realm for embodied agents to explore, which requires the acquisition of various basic skills (e.g., crafting logs) and the ability to plan and execute diverse tasks.
First, as shown in Figure~\ref{fig:intro} (a), the LLM-based agent produces uncontrollable outputs. 
The success of the agent's responses hinges heavily on careful prompt engineering~\citep{huang2022inner}, ensuring that the LLM comprehends the environment and task objectives. 
Moreover, a universally applicable prompt that suits every LLM and task is an unattainable goal.
Therefore, this prompting process is labor-intensive and contradicts our aim of enabling agents to act in a self-driven manner.
Second, when compared to visual feedback, language often encounters difficulties in intuitively conveying specific world concepts (e.g., recipes) to users, as illustrated in Figure~\ref{fig:intro} (b), thereby unavoidably creating obstacles for robust human-computer/AI interaction~\citep{preece1994human,fallman2003design}.

Unlike LLMs, humans possess an innate ability to process and generate information through both visual and text channels.
This inherent gift significantly enhances our capability to interact with the world. 
However, the coupling of LLM-based agents with multimodal I/O has been relatively underexplored in an open-ended environment.
To fill this gap, we introduce \textbf{Steve-Eye} \includegraphics[height=10pt]{icons/eye.png}, a large multimodal model that enables LLM-based embodied agents to engage with the open world via visual-text interfaces.
Steve-Eye excels at producing responses that demonstrate a comprehensive grasp of the environment, common-sense reasoning, and executable skill plans.
To achieve this, Steve-Eye is equipped with three indispensable functions: (1) multimodal perception; (2) foundational knowledge base; and (3) skill prediction and planning. 
In this paper, we choose Minecraft as our validation platform considering its vast sandbox world and the high degree of freedom.
More environments can also be considered, e.g., Virtual Home~\citep{puig2018virtualhome}, AI2THOR~\citep{kolve2017ai2}.
Due to the space limit, we discuss the exploration of more generic environments in Appendix~\ref{sec:sup_env} and leave it as our future work.
Our contributions can be summarized as follows:

\noindent\textbf{Open-World Instruction Dataset. }
We construct an extensive instruction dataset to train Steve-Eye for the acquisition of three mentioned functions.
The instruction data contains not only the agent's per-step status and environmental features but also the essential knowledge for agents to act and plan.
However, collecting such a dataset in an open world can be a costly endeavor, especially when aiming to gather fine-grained and diverse labels.
As a result, previous studies~\citep{fan2022minedojo} have often relied on readily available unsupervised data (e.g., video-subtitle pairs) for pre-training.
In these approaches, the agent's comprehension of its status and environment is implicitly learned through self-supervised techniques, while its foundational knowledge is directly derived from general-purpose LLMs.
In contrast, our work involves curating multimodal instructional data specifically designed for open-ended embodied agents, by utilizing ChatGPT~\citep{chatgpt}.


\textbf{Large Multimodal Model and Training. }
Steve-Eye combines a visual encoder which converts visual inputs into a sequence of embeddings, along with a pre-trained LLM which empowers embodied agents to engage in skill or task reasoning in an open world.
During the training process, we employ a two-stage strategy similar to ~\citet{liu2023visual}.
This strategy commences with the alignment of multimodal elements between the visual encoder and the large language model, followed by the instruction tuning through our constructed dataset.

\textbf{Open-World Benchmarks. }
We carry out extensive experiments to demonstrate that our proposed Steve-Eye outperforms LLM-based agents in open-world setups. 
Specifically, we develop the following benchmarks to evaluate agent performance from a broad range of perspectives:
(1) environmental visual captioning (ENV-VC), which assesses an agent's capacity to perceive and describe its surroundings effectively; 
(2) foundational knowledge question answering (FK-QA), which evaluates the proficiency in mastering basic knowledge crucial for an agent's decision-making;
(3) skill prediction and planning (SPP), which quantifies an agent's capability to act and plan strategically.

%% file: sections/2_relatedwork.tex
\section{Related Work}

\subsection{Open-world Embodied Agents with LLMs}
The rapid progress of large language models~\citep{brown2020language,raffel2020exploring,zhang2022opt,chowdhery2022palm} has significantly boosted their capacity to encode a wide range of human behaviors within training data~\citep{bommasani2021opportunities}.
When equipped with narrowly designed prompts, LLM-based agents exhibit the capability to generate executable plans for tasks such as indoor robot manipulation.
For instance, SayCan~\citep{ahn2022can} integrates skill affordances with LLMs to yield actionable plans, while Palm-E~\citep{driess2023palm} takes a step further by constructing hierarchical agents capable of handling multimodal prompts. 
This approach has also proven its efficacy in open-world environments~\citep{huang2022language,li2022pre}.
In contrast to robot manipulation, agents in the wild require a heightened level of real-time situational awareness and foundational knowledge to execute intricate skill plans across a diverse array of tasks.
To simulate human behaviors in such open worlds, Generative Agents~\citep{park2023generative} store agents' experiences and retrieve these memories to generate plans in a text-based sandbox game.

In recent years, the 3D sandbox Minecraft has received considerable attention owing to its remarkably flexible game mechanics to serve as a prominent open-world benchmark (e.g., MineRL~\citep{guss2019minerl} and Minedojo~\citep{fan2022minedojo}).
DEPS~\citep{wang2023describe} introduces the descriptor, explainer, and selector for plan generation with the help of LLM.
Plan4MC~\citep{yuan2023plan4mc} constructs a skill graph and proposes a skill search algorithm to minimize planning errors. 
Voyager~\citep{wang2023voyager} proposes an LLM-powered lifelong learning agent that continually explores the Minecraft world. 
Similar to \citep{park2023generative}, GITM~\citep{zhu2023ghost} integrates LLMs with text-based memory and knowledge to create generic agents in Minecraft.
Among these studies, Voyager~\citep{wang2023voyager} and GITM~\citep{zhu2023ghost} lean entirely on text descriptions of the environment to act and plan, while Plan4MC ~\citep{yuan2023plan4mc} and DEPS \citep{wang2023describe} have visual-input skills but still rely on merely text for planning. 
None of them try to understand the rich visual observation provided natively by Minecraft. 
In contrast to these works, our work trains a large multimodal model to fill this gap.

\subsection{Large Multimodal Models (LMMs)}
In comparison to LLMs, large multimodal models (LMMs)~\citep{awadalla2023openflamingo} encompass a broad range of information beyond text modality, which can be categorized into two primary streams. 
The first category~\citep{gupta2023visual,huang2023audiogpt,patil2023gorilla,suris2023vipergpt} 
involves hinging on ChatGPT~\citep{chatgpt} or GPT-4~\citep{gpt4} to generate in-context responses without parameter tuning.
However, these approaches heavily rely on the availability of an LLM's API and the quality of the designed prompts.
The second category comprises end-to-end pre-trained models. 
Within this category, models such as~\citet{huang2023language,peng2023kosmos} are trained entirely from scratch.
Conversely, some research explores efficient fine-tuning using pre-trained LLMs by incorporating lightweight modality encoders, such as Qformer~\citep{li2023blip} or Perceiver~\citep{alayrac2022flamingo}. Recently, \citet{liu2023visual} propose to explicitly instruction-tune a LLM using vision-language instruction data.

In this work, we propose Steve-Eye by building upon pre-trained LLMs, aiming to develop an open-world agent powered by a large-scale model with versatile multimodal I/O capabilities.

%% file: sections/3_method.tex
\section{Methodology}
In this section, we first provide our instruction-following dataset to develop three key functions for the agent's open-world interaction in Section~\ref{sec:data}.
We then propose our large multimodal agent Steve-Eye in Section~\ref{sec:model}, and clarify details of the training procedure in Section~\ref{sec:train}.
We adopt Minecraft as our open-ended platform in this paper to collect data and validate the model, anticipating to explore a broader range of environments for Steve-Eye in future studies.

To empower an agent with the self-driven capacity to act and plan in an open world, we posit that the following embodied functions are indispensable:
(1) multimodal perception function which offers a detailed description of the agent status and environmental features; 
(2) foundational knowledge base which imparts an understanding of how the world works and conveys crucial basic knowledge related to skills and tasks;
(3) skill prediction and planning which is responsible for generating skill execution feedback (e.g., success or failure) and crafting high-level skill plans for handling more complex and long-horizon tasks.
We develop these functions by building the corresponding instruction dataset to pre-train Steve-Eye as follows.

\subsection{Open-World Instruction-Following Dataset}
\label{sec:data}

\begin{wrapfigure}{r}{0.35\textwidth} 
\vspace{-0.3cm}
\centering
  \includegraphics[width=0.35\textwidth]{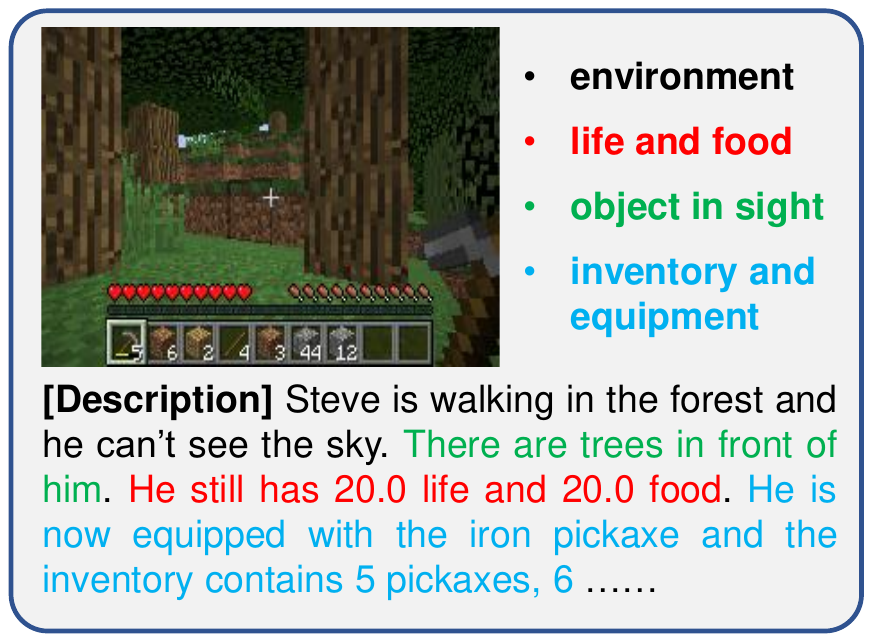} 
  \vspace{-6mm}
  \caption{Multimodal perception}
\label{fig:mm_data1} 
\vspace{-0.3cm}
\end{wrapfigure}

\noindent\textbf{Multimodal Perception Instructions. }
Human players can perform actions in Minecraft mainly relying on their visual perception, without any prior hints or imposed game judgments.
In order to endow Steve-Eye with the same ability, it is required to provide it with comprehensive visual descriptions of the environment.
To achieve this, we use Minedojo~\citep{fan2022minedojo} to obtain Minecraft snapshots which contain a wide array of details within the agent's surroundings, including environmental features, the agent's life and food status, inventory items, and equipment, as illustrated in Figure~\ref{fig:mm_data1}.
In addition, we leverage MaskCLIP~\citep{zhou2022extract} to identify the in-sight objects of these snapshots without supervised annotations.
During our data collection process, for each snapshot $\mathcal{I}$ and its corresponding description $\mathcal{X}_{C}$, we initiate a three-step approach.
Firstly, we prompt ChatGPT to curate a list of 40 instructions as shown in Figure~\ref{fig:description_inst} in Appendix~\ref{sec:sup_data_mm}.
Then we enrich snapshot details as dense caption to describe its content, with the assistance of ChatGPT.
Finally, we select an instruction $\mathcal{X}_{Q}$ randomly from the list and combine it with the snapshot's caption to create a single-round multimodal description pair (e.g., \#\#\# Human: $\mathcal{X}_{Q}$ $\mathcal{I}$\textbackslash n \#\#\# Embodied Agent: $\mathcal{X}_{C}$\textbackslash n.).
By doing so, we collect 200K instructional pairs for multimodal perception learning.

\begin{wrapfigure}{r}{0.36\textwidth} 
\vspace{-0.3cm}
\centering
  \includegraphics[width=0.35\textwidth]{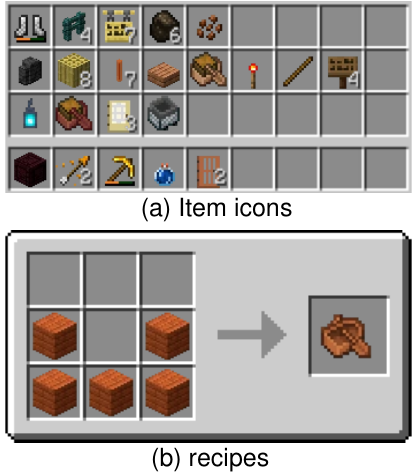} 
  \vspace{-2mm}
  \caption{Icons and recipes}
\vspace{-0.3cm}
\label{fig:icons} 
\end{wrapfigure}

\noindent\textbf{Foundational Knowledge Instructions. }
Embodied agents require a foundation of essential knowledge to facilitate action-taking and skill planning.
In Minecraft, such knowledge should contain item recipes, details of item attributes, their associated numerical value, etc.
We access this vital information from Minecraft-Wiki~\citep{mcwiki}, which comprises an extensive collection of over 9,000 HTML pages.
To be specific, we first obtain all item icons from Minecraft-Wiki and generate 200K icon inventory images, as illustrated in Figure~\ref{fig:icons} (a).
Each icon image corresponds to a 4-row table with an associated caption adhering to a standardized template: ``There is a Minecraft inventory with 4 rows. From left to right, they are ...''.
As shown in Figuire~\ref{fig:icon_inst} in Appendix~\ref{sec:sup_data_fk}, we curate a set of 20 distinct prompts designed to challenge the model's ability to recognize items.
Subsequently, we further collect all recipe-related information from the Wiki as illustrated in Figure~\ref{fig:icons} (b), and design similar prompt templates to formulate 10,000 recipe-image instructional pairs.
Lastly, we process the Wiki and utilize this corpus to produce 40,000 single-round question-answer pairs.
In total, we collect a high-quality dataset with 250K foundational knowledge instructions.

\noindent\textbf{Skill-related Interaction Instructions. }
The environmental description and foundational knowledge serve as prerequisites for an agent's interaction within the open world.
However, a successful interaction requires more than these elements alone. 
It relies upon the mastery of basic skills, such as log, harvesting, and food preparation, as well as high-level skill planning abilities to tackle complex, long-horizon tasks, such as crafting an iron pickaxe.
To facilitate this, we gather corresponding training data for skill prediction and planning, which enables our model to provide correct feedback on both basic skills and long-horizon tasks across a spectrum of agent or environmental conditions.
Specifically, the data collection process involves two steps. 
First, we sample skill trajectories based on the pre-trained basic skill policies and collect 200K snapshot pairs with corresponding statuses from these trajectories.
Each snapshot pair $\{\mathcal{I}_0, \mathcal{I}_t\}$ denotes the 0-th and t-th timestamp of the skill trajectory.
Next, we employ ChatGPT to generate question-answer pairs about diverse aspects of skill execution status.
These questions delve into whether the agent completes the skill, encounters unexpected failures, or seeks explanations for such failures.
More details can be found in Appendix~\ref{sec:sup_data_spp}.
Second, we sample 40K task trajectories using the planner in~\citet{yuan2023plan4mc}, each of which can be denoted as $\mathcal{T}=\{s_1, s_2,...s_{\rm T}\}$ representing the task is finished via a ${\rm T}$-round planning procedure, where $s_i$ is the skill plan for $i$-th round.
At each round $i$, we feed our model with its start snapshot and task initialization, and curate instructional questions to inquire about $s_i$ with reasonable explanation.
In this manner, we obtain 200K instructional pairs from task trajectories.

\begin{figure}[!t]
\centering
\includegraphics[width=0.9\textwidth]{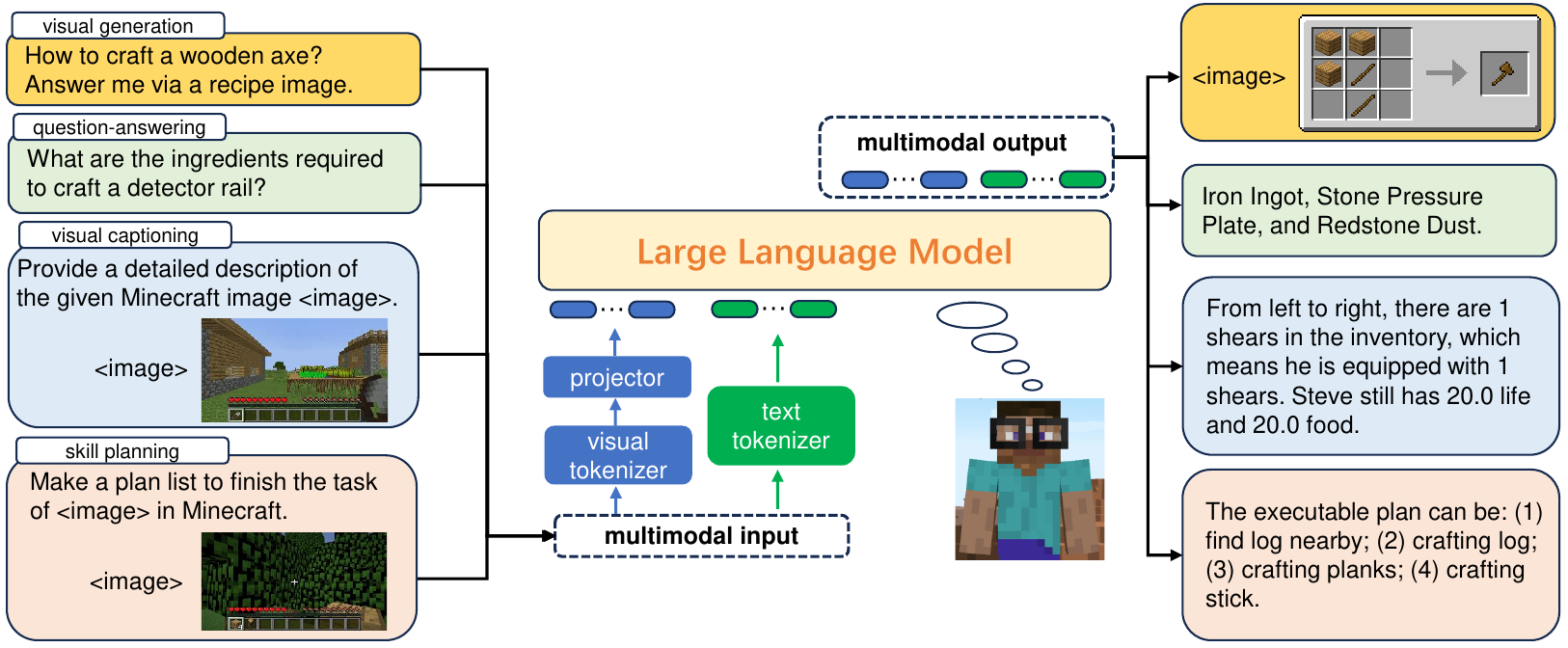}
\caption{
Illustration of Steve-Eye: a large multimodal model designed to seamlessly process both visual and language inputs.
Steve-Eye excels in acquiring fundamental knowledge of the world it lives in, 
understanding the nuances of its surroundings, and generating executable plans to complete a wide array of open-ended tasks.
Furthermore, Steve-Eye responds to user instructions through either visual or text-based cues, enhancing the convenience and flexibility of human-AI interaction.
}
\label{fig:method}
\end{figure}

\subsection{Model Architecture}
\label{sec:model}
Figure~\ref{fig:method} illustrates the overall architecture of our proposed model.
Steve-Eye, functioning as a generative model, connects an image-oriented tokenizer $f_{v}$ with the pre-trained LLM backbone $\Theta$.
We adopt the image tokenizer, e.g., VQ-GAN~\citep{esser2021taming}, to encode the raw images $\mathcal{I}$ into token embeddings $\mathcal{V}=\{v_1,v_2,...,v_n\} \in \mathbbm{R}^{n \times d}$, where $n$ denotes the number of visual tokens and $d$ is the dimensionality of each token.
We further utilize a lightweight projection module $f_l$ with a trainable projection matrix $W$.
This module maps the visual tokens to the same space with text embeddings, yielding $\hat{\mathcal{V}}=\{\hat{v}_1,\hat{v}_2,...,\hat{v}_n\} \in \mathbbm{R}^{n \times \hat{d}}$:
\begin{equation}
\hat{\mathcal{V}} = W \mathcal{V}; \hspace{0.3em} \text{where} \hspace{0.3em}  \mathcal{V} = f_v(I).
\end{equation}

To effectively process visual-language inputs and generate corresponding outputs, our model integrates the visual codebook $\mathcal{C}_v$ into the pre-existing language vocabulary $\mathcal{C}_l$.
This integration leads to the formation of a unified multimodal codebook, denoted as $\mathcal{C}_m=\mathcal{C}_v \cup \mathcal{C}_l$.
Additionally, in order to mark the starting and ending points of visual elements in I/O sequences, we introduce two special tokens, namely $<$vis$>$ and $<$/vis$>$.
The LLM backbone $\Theta$ of our Steve-Eye is built upon a decoder-only architecture with casual transformers.
Our model employs an auto-regressive prediction mechanism, generating responses based on the provided multimodal input tokens.
The resulting response is a mixed sequence of visual and textual tokens, represented as $\mathcal{Y}=\{y_1,y_2,...,y_m\}$.
For each embedding $y_i$, we pass it through a linear layer $f_p$ followed by a softmax operation, mapping it into a probability distribution of the multimodal vocabulary.
The final prediction for the $i$-th token $z_i$ is determined by selecting the token from the multimodal codebook with the highest score:
\begin{equation}
z_i = \argmax( \text{softmax}(f_p(y_i))).
\end{equation}

\subsection{Training}
\label{sec:train}
Each instruction-following instance can be formulated as a multi-round conversation $\{\mathcal{X}_Q^1, \mathcal{X}_C^1,...,$ $\mathcal{X}_Q^N, \mathcal{X}_C^N\}$, where each $\{\mathcal{X}_Q^i, \mathcal{X}_C^i\}$ represents a question-answer interaction between a human and the embodied agent and $N$ indicates the total number of rounds in the conversation.
The entire instructional dataset follows this unified template, as demonstrated in Figure~\ref{fig:uni_prompt} in Appendix~\ref{sec:sup_data_train}.
To efficiently train our model, we employ the negative log-likelihood objective over the prediction tokens with instruction tuning:
\begin{equation}
    \mathcal{L}(\Theta)=-\sum_{j=1}^{L} \log P_{\Theta}(y_j|\mathcal{I}, \hat{y}_{1:j-1}),
\end{equation}
where $y$ and $\hat{y}$ respectively denote the input and target token sequences, with $\Theta$ representing the model parameters, and $L$ representing the length of the target sequence.
The input visual content $\mathcal{I}$ may represent an empty image depending on the input instruction.
It is worth noting that we constrain the loss computation to only consider the answer tokens $\mathcal{X}_C$.
This constraint prevents training from becoming excessively straightforward and ensures that the model's primary focus is on learning to precisely generate coherent responses.
Similar to~\citet{liu2023visual}, we adopt a two-stage instruction-tuning strategy to train our model:

\noindent\textbf{Two-Stage Instruction-Tuning.}
\textbf{(1) Multimodal feature alignment}: 
In the first stage, our primary objective is to align visual features with the language token space.
In order to strike a balance between efficient tuning and a comprehensive coverage of the world's concepts, we curate our open-ended instruction dataset to 600K snapshot-text pairs. 
These pairs are then transformed into instruction-following data as described in Section~\ref{sec:data}.
During the feature alignment stage, we maintain the visual encoder and the LLM parameters in a frozen state, exclusively training the projection module.
Additionally, this training phase involves fine-tuning token embeddings to accommodate the newly introduced visual codebook and two special tokens $<$vis$>$ and $<$/vis$>$.
\textbf{(2) End-to-end instruction tuning}: 
In the second stage, we continue to keep the visual encoder frozen while concurrently training the projection module and LLM.
This second stage leverages the entire open-ended instructions and contributes significantly to enhancing the model's capability of comprehending and effectively responding to complex multimodal instructions.

%% file: sections/4_experiments.tex
\section{Experiments}
\subsection{Experimental Setup}
\noindent\textbf{Implementation Details. } 
In this paper, we use the LLaMA-2 model~\citep{touvron2023llama2} as the LLM backbone.
Additionally, we use CLIP~\citep{radford2021learning} as our visual encoder to achieve the best performance for non-visual generative tasks, and use VQ-GAN~\citep{esser2021taming} as the default visual tokenizer for visual generation.
The size of visual codebook  $\mathcal{C}_v$ and language vocabulary is 8192 and 32000, respectively. 
In addition, we add $<$vis$>$ and $<$/vis$>$ to the final unified codebook, indicating the starting and ending points of visual content.
Similar to~\citet{liu2023visual}, we construct 850K instruction-answer pairs for model training.
Note that the model is trained to predict the agent's answer, and thus only sequence/tokens of answer will be used to compute the loss in the auto-regressive model.
We also adopt LoRA~\citep{hu2021lora} to reduce the computational cost for efficient tuning.
We choose MineDojo~\citep{fan2022minedojo} as the Minecraft platform to collect our instruction data and conduct experiments.  
Following \citet{yuan2023plan4mc}, we use the environments of programmatic tasks to train basic policies with RL.
These policies are trained to execute corresponding skills and keep fixed in all testing tasks.

\noindent\textbf{Evaluation Benchmarks. }
We conduct experiments on three benchmarks to evaluate an agent's interaction ability in an open world.
\textbf{(1) Environmental visual captioning (ENV-VC)}: given a snapshot, the model is asked to describe the agent's current status and environmental features from diverse aspects (e.g., life, food...).
We evaluate the prediction's accuracy of each aspect by extracting corresponding answers from the output description to compare with the groundtruth.
\textbf{(2) Foundational knowledge question answering (FK-QA)}: to assess the model's grasp of essential knowledge, we collect a set of 10,000 Minecraft-related questions from different sources, including the Wiki pages, Wiki tables, and Minecraft recipes.
The performance is measured by the model's ability to provide correct answers to these questions.
\textbf{(3) Skill prediction and planning (SPP)}: we utilize our proposed Steve-Eye to predict whether a skill has been successfully completed and assert its capability to generate executable high-level skill plans for long-horizon tasks.

\subsection{Environmental Visual Captioning (ENV-VC)}
We introduce this evaluation protocol for asserting Steve-Eye’s multimodal perception function, which serves as an initial stride toward comprehensive evaluation of large multimodal models.
Specifically, we collect 20,000 Minecraft snapshots (named ENV-VC test set) using Minedojo and apply the proposed data generation pipeline to create six questions for each snapshot, resulting in a total of 120K questions.
These six questions pertain to the prediction of various aspects, including inventory items~\includegraphics[height=8pt]{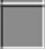}, equipment~\includegraphics[height=8pt]{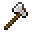}, objects in sight~\includegraphics[height=8pt]{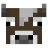}, life~\includegraphics[height=8pt]{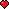}, food~\includegraphics[height=8pt]{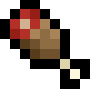}, and the visibility of sky~\includegraphics[height=8pt]{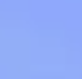}.

During the inference phase, 
Steve-Eye predicts answers based on these questions and the input snapshot.
Experimental results are presented in Table~\ref{tab:env-vc} and Table~\ref{tab:env-vc-1}.
As shown in Table~\ref{tab:env-vc}, our visual encoder, when combined with multimodal instruction tuning, significantly enables the ability of the text-only language model LLM (Llama-2-7b) to comprehend the contents of the snapshots (Steve-Eye-7b).
Notably, Steve-Eye outperforms BLIP-2 by a substantial margin due to the improved reasoning ability enabled by the larger LLM.
Furthermore, the visual encoder plays a crucial role in facilitating multimodal understanding. 
Surprisingly, the model equipped with CLIP~\citep{radford2021learning} surpasses the performance of the model using MineCLIP~\citep{fan2022minedojo}, achieving over \expbf{48.9}, \expbf{21.0} and \expbf{19.9} improvements in inventory, equipment, and object-in-sight predictions, respectively.
We attribute this performance difference to the fact that MineCLIP does not prioritize fine-grained alignment during pre-training, despite being exposed to a diverse range of Minecraft videos.
In summary, Steve-Eye's ability to comprehend visual cues from its surroundings lays the foundation for subsequent interactions with the world.

\input{tables/env-vc-1}

To investigate the effictiveness of various types of instructional data for multimodal perception, we carry out experimental comparisons with diverse data configurations in Table~\ref{tab:env-vc-1}.
First, our results showcase a significant improvement in the model's capacity to respond to instructional questions through instruction tuning, which leads to impressive gains of over ~\expbf{50} for inventory, equipment, and object-in-sight prediction.
Furthermore, the inclusion of the multimodal perception dataset and icon images in the training data both contribute to a substantial improvement in the model's overall performance.
Ultimately, the best results are achieved when combining all available data sources.

\input{tables/env-vc-2}

\subsection{Foudational Knowledge Question Answering (FK-QA)}
Following ~\citet{vicuna}, we establish a question database specialized to assess our model's proficiency in generating responses pertaining to fundamental Minecraft knowledge. 
This evaluation is carried out through a validation dataset known as the FK-QA test set, which is further divided into two distinct subsets: TEXT and IMG.
In the FK-QA TEXT subset, we generate a collection of 10,000 question-answer pairs curated from various sources, including the Minecraft-Wiki pages, Minecraft-Wiki tables, and Minecraft recipes.
Each category comprises 2,000, 5,000, and 3,000 pairs, respectively. 
Upon receiving a response from Steve-Eye, we feed both the generated response and the corresponding groundtruth answer to ChatGPT.
ChatGPT will first examine the accuracy of the response as a measure of answer correctness.
To minimize variability in error, ChatGPT conducts a further evaluation, considering the response's accuracy, relevance, and level of detail. 
This comprehensive evaluation yields an overall score on a scale ranging from 0 to 10, where a higher score signifies superior overall performance.
In the FK-QA IMG subset, we shift our focus to visual generation by employing 3,000 recipe images as groundtruth data. 
Here, our model is tasked with generating visual outputs for each item within the recipe inventory, following a specific order.
The visual output is considered correct only if every element of the recipe is accurately generated.
We adopt this metric to assert our model's ability to produce multimodal feedback.

Table~\ref{tab:fk-qa} presents both scoring and accuracy results.
It's worthy to note that Llama-2 exhibits consistent performance regardless of the model's scale, with Llama-2-70b only marginally outperforming the 7b-version by \expbf{1.26} in accuracy, meanwhile 13b-version performs even worse than 7b-version on the scoring results.
We hypothesize that this phenomenon can be attributed to distinct variations in difficulty levels encountered within our FK-QA test set.
Llama-2 fails to answer correctly for the challenging part regardless of its size due to essential knowledge missing.
In contrast, Steve-Eye outperforms both Llama-2 and gpt-turbo-3.5, despite its considerably smaller scale.
Furthermore, our model exhibits a more substantial improvement in responding to Recipe and Wiki Table questions as compared to Wiki Page questions. 
This disparity can likely be attributed to the fact that Wiki Page contains a large proportion of invalid questions (e.g., version, history), whereas Recipe and Wiki Table predominantly feature knowledge-related inquiries.
Such result further validates the effectiveness of our approach in acquiring foundational knowledge.
Unlike text-only LLMs, our model exhibits considerable ability to output visual contents, which achieves $65.13\%$ accuracy on FK-QA IMG using the 13b-version.
The multimodal generation ability enables Steve-Eye to better serve as an assistant for potential needed people such as beginners of this game.
We show more details and cases in Appendix~\ref{sec:sup_mm_gen}.

\input{tables/fk-qa}

\subsection{Skill Prediction and Planning (SPP)}

\noindent\textbf{Skill Prediction. } 
Similar to Section~\ref{sec:data}, we collect another 20K snapshot pairs in the form of $\{\mathcal{I}_0, \mathcal{I}_t\}$ from skill trajectories (referred to as Skill-Pred test).
These pairs are input into our model to query the current execution status of the skill.
The execution status can fall into one of three categories: success, failure, and running, with ``running'' signifying that the skill is currently in progress.
\input{tables/skill-pred} 
As shown in Table~\ref{tab:skill-pred}, our model exhibits commendable performance in skill status prediction.
However, the performance is still far from enough to completely replace the rule-based game judgment adopted by the existing RL-based skill agents.
These experiments indicate that, despite the excellent multimodal understanding capabilities of our model in open-world environments in previous experiments, it still falls short in fine-grained reasoning tasks that involve consecutive frames to some extent.

\noindent\textbf{Skill Planning. }
Following~\citet{yuan2023plan4mc}, we carry out evaluation on 24 difficult tasks in Minecraft.
These tasks can be categorized into three types: 
cutting trees to craft primary items (7), mining cobblestones to craft advanced items (7), and interacting with mobs to harvest food and materials (10).
Each task is tested for 30 episodes, where an episode refers to a multi-round interaction process.
At each round, the model receives the environmental feedback from the last round, plans a skill list based on the current status, and then picks up the top skill to execute.
For each task episode, we set a maximum step between [3000, 10000].
In our evaluation, we compare Steve-Eye against two baseline approaches: (1) MineAgent~\citep{fan2022minedojo}, which completes tasks without decomposing them into basic skills, and uses PPO and self-imitation learning with CLIP reward, and (2) GPT Assistant, which employs ChatGPT as a high-level planner to generate skill plans by prompting itself with information from the environment and the agent's status.
The results in Table~\ref{tab:skill_plan} demonstrate that Steve-Eye significantly outperforms both baseline methods.
Additionally, we conduct experiments in which Steve-Eye takes over the skill prediction function from the rule-based game judgment in Minecraft.
This self-driven variant is referred to as `Steve-Eye-auto.' 
Since the model's skill prediction is not always 100\% accurate, Steve-Eye-auto does experience some performance degradation when compared to Steve-Eye.
This degradation is more pronounced in longer, complex tasks (e.g.,\mcstonepickaxe, \mcfurnace, \mccobblestonewall) as opposed to short-term tasks (e.g.,\mccraftingtable, \mcbowl,\mcbed).
Nevertheless, Steve-Eye-auto still demonstrates significant performance improvements in most tasks, compared to the baselines.
For additional details about this benchmark, please refer to Appendix~\ref{sec:sup_plan}.


\input{tables/skill-plan}

\begin{figure}[htb]
\centering
\includegraphics[width=0.7\textwidth]{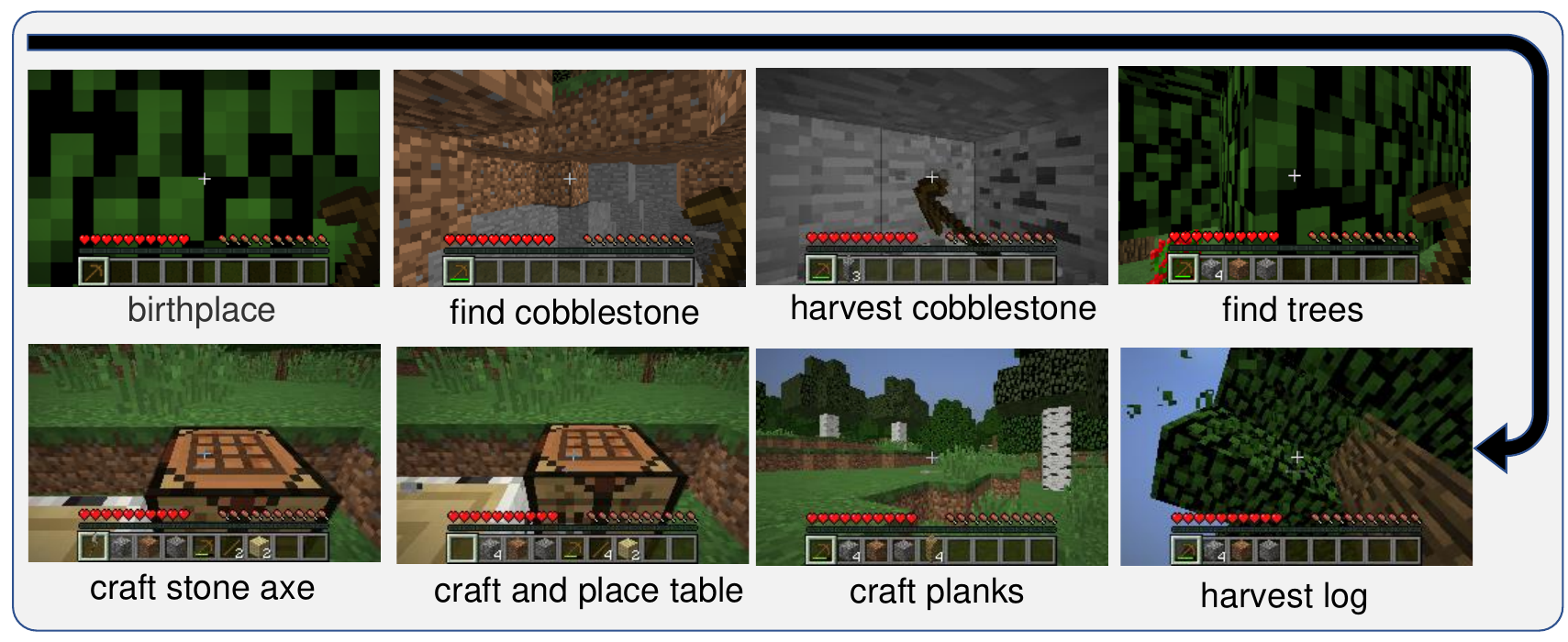}
\caption{
Snapshots of a qualitative example, illustrating how Steve-Eye completes the task of ``crafting a stone axe with a wooden pickaxe." 
Our model generates a skill plan at each interaction round and selects the top skill from the plan list for execution.
}
\label{fig:vis}
\end{figure}

For better visualization, we provide a qualitative example of Steve-Eye completing the task ``crafting stone axe with wooden pickaxe'' as shown in Figure~\ref{fig:vis}.

%% file: tables/env-vc-1.tex
\begin{table}[!t]
\small
\centering
\caption{Comparisons of different model settings on the environmental visual caption benchmark. The experiments are conducted on 20K ENV-VC test set.}
\vspace{-2mm}
\setlength{\tabcolsep}{5pt}
\scalebox{0.9}{
\begin{tabular}{l|ccccccc}
\toprule
Model & visual encoder & 
inventory \includegraphics[height=8pt]{icons/inventory.png} & 
equip~\includegraphics[height=8pt]{icons/iron-axe.png} & 
object in sight~\includegraphics[height=8pt]{icons/cow.png} & 
life~\includegraphics[height=8pt]{icons/heart.jpg} & 
food~\includegraphics[height=8pt]{icons/hunger.png} & 
sky~\includegraphics[height=8pt]{icons/sky.png} \\
\midrule
BLIP-2 & CLIP & 41.6 & 58.5 & 64.7 & 88.5 & 87.9 & 57.6 \\
Llama-2-7b & - & - & - & - & - & - & - \\
Steve-Eye-7b & VQ-GAN & 89.9 & 78.3 & 87.4 & 92.1 & 90.2 & 68.5 \\
Steve-Eye-13b & MineCLIP & 44.5 & 61.8 & 72.2 & 89.2 & 88.6 & 68.2 \\
Steve-Eye-13b & VQ-GAN & 91.1 & 79.6 & 89.8 & 92.7 & 90.8 & 72.7 \\
Steve-Eye-13b & CLIP & \textbf{92.5} & \textbf{82.8} & \textbf{92.1} & \textbf{93.1} & \textbf{91.5} & \textbf{73.8} \\
\bottomrule
\end{tabular}
}
\label{tab:env-vc}
\end{table}

%% file: tables/env-vc-2.tex
\begin{table}[!t]
\centering
\caption{Comparisons of different data configurations on the environmental visual captioning benchmark, where ``snapshot desc.'' denotes the 200K multimodal perception instruction dataset.}
\vspace{-2mm}
\small
\setlength{\tabcolsep}{5pt}
\scalebox{0.9}{
\begin{tabular}{l|llllll}
\toprule
  & 
inventory \includegraphics[height=8pt]{icons/inventory.png} & 
equip~\includegraphics[height=8pt]{icons/iron-axe.png} & 
object in sight~\includegraphics[height=8pt]{icons/cow.png} & 
life~\includegraphics[height=8pt]{icons/heart.jpg} & 
food~\includegraphics[height=8pt]{icons/hunger.png} & 
sky~\includegraphics[height=8pt]{icons/sky.png} \\
\midrule
no instruction tuning & 22.7 & 24.3 & 39.8 & 81.2 & 80.4 & 61.1 \\
w/o snapshot desc. & 46.2 \tiny{\textcolor{red}{(+23.5)}} &  40.9 \tiny{\textcolor{red}{(+16.6)}} & 41.2 \tiny{\textcolor{red}{(+1.4)}} & 83.0 \tiny{\textcolor{red}{(+1.8)}} & 82.4 \tiny{\textcolor{red}{(+2.0)}} & 63.3 \tiny{\textcolor{red}{(+2.1)}}  \\
w/o icon images & 52.3 \tiny{\textcolor{red}{(+29.6)}} & 48.1 \tiny{\textcolor{red}{(+23.8)}} & 91.4 \tiny{\textcolor{red}{(+51.6)}} & 92.5 \tiny{\textcolor{red}{(+11.3)}} & 90.9 \tiny{\textcolor{red}{(+10.5)}} & 73.5 \tiny{\textcolor{red}{(+12.4)}} \\
full data & 92.5 \tiny{\textcolor{red}{(+69.8)}} & 82.8 \tiny{\textcolor{red}{(+58.5)}} & 92.1 \tiny{\textcolor{red}{(+52.3)}} & 93.1 \tiny{\textcolor{red}{(+11.9)}} & 91.5 \tiny{\textcolor{red}{(+11.1)}} & 73.8 \tiny{\textcolor{red}{(+12.7)}}\\
\bottomrule
\end{tabular}
}
\label{tab:env-vc-1}
\end{table}

%% file: tables/fk-qa.tex
\begin{table}[!t]
\centering
\caption{Comparisons on FK-QA test set of the foundational knowledge question answering benchmark. The evaluation metrics consider both the scoring and accuracy dimensions simultaneously.}
\vspace{-2mm}
\small
\scalebox{0.9}{
\begin{tabular}{p{2cm}|p{1.5cm}p{1.5cm}p{1.5cm}p{1.8cm}|p{1.4cm}c}
\toprule
\multirow{2}{*}{} & \multicolumn{4}{c}{Scoring} & \multicolumn{2}{c}{Accuracy} \\
\cmidrule{2-7}
& Wiki Page & Wiki Table &  Recipe & TEXT All & TEXT & IMG \\
\midrule
Llama-2-7b  & 6.90 & 6.21 & 7.10 & 6.62 & 37.01\% & - \\
Llama-2-13b & 6.31 \tiny{(-0.59)} & 6.16 \tiny{(-0.05)} & 6.31 \tiny{(-0.79)}  & 6.24 \tiny{(-0.38)} & 37.96\% & - \\
Llama-2-70b & 6.91 \tiny{\textcolor{red}{(+0.01)}} & 6.97 \tiny{\textcolor{red}{(+0.76)}} & 7.23 \tiny{\textcolor{red}{(+0.13)}} & 7.04 \tiny{\textcolor{red}{(+0.42)}} & 38.27\% & - \\
gpt-turbo-3.5 & 7.26 \tiny{\textcolor{red}{(+0.36)}} & 7.15 \tiny{\textcolor{red}{(+0.94)}} & \textbf{7.97} \tiny{\textcolor{red}{(+0.87)}} & 7.42 \tiny{\textcolor{red}{(+0.80)}} & 41.78\% & - \\
Steve-Eye-7b & 7.21 \tiny{\textcolor{red}{(+0.31)}} & 7.28 \tiny{\textcolor{red}{(+1.07)}} & 7.82 \tiny{\textcolor{red}{(+0.72)}} & \textbf{7.54} \tiny{\textcolor{red}{(+0.92)}} & 43.25\% & 62.83\% \\
Steve-Eye-13b & \textbf{7.38} \tiny{\textcolor{red}{(+0.48)}} & \textbf{7.44} \tiny{\textcolor{red}{(+1.23)}} & 7.93 \tiny{\textcolor{red}{(+0.83)}} & \textbf{7.68} \tiny{\textcolor{red}{(+1.06)}} & \textbf{44.36\%} & \textbf{65.13\%} \\
\bottomrule
\end{tabular}
}
\vspace{-0.2mm}
\label{tab:fk-qa}
\end{table}

%% file: tables/skill-pred.tex
\begin{wraptable}{r}{0.5\textwidth}
\vspace{-1mm}
\centering
\caption{Recall/Accuracy results on Skill-Pred test set for the skill prediction benchmark.}
\small
\vspace{-3mm}
\setlength{\tabcolsep}{2.2pt}
\begin{tabular}{lccc}
\toprule
 & running (\%) & success (\%) & fail (\%) \\
\midrule
BLIP-2 & 65.2/58.8 & 49.8/54.3 & 42.1/51.8 \\
Steve-Eye-7b & 89.8/82.5 & 77.6/81.4 & 74.2/79.9 \\
Steve-Eye-13b & 92.1/84.2 & 80.5/83.1 & 76.8/81.5 \\
\bottomrule
\label{tab:skill-pred}
\end{tabular}
\vspace{-3mm}
\end{wraptable}

%% file: tables/skill-plan.tex
\begin{table}[!t]
\centering
\caption{Comparisons on the skill planning benchmark. We test the mean success rates of all tasks, where each task is executed for 30 episodes using the same seeds for initialization. }
\centering
\vspace{-2mm}
\begin{subtable}{1\linewidth}
\centering
   \scalebox{0.8}{
    \begin{tabular}{l|ccccccc|ccccccc}
    \toprule
    Model & \mcstick & \mccraftingtable & \mcbowl & \mcchest & \mctrapdoor & \mcsign 
          & \mcwoodenpickaxe & \mcfurnace & \mcstonestairs & \mcstoneslab & \mccobblestonewall & \mclever & \mctorch & \mcstonepickaxe \\
    \midrule
    MineAgent & 0.00 & 0.03 & 0.00 & 0.00 & 0.00 & 0.00 & 0.00 & 0.00 & 0.00 & 0.00 & 0.21 & 0.0 & 0.05 & 0.0 \\
    gpt assistant & 0.30 & 0.17 & 0.07 & 0.00 & 0.03 & 0.00 & 0.20 & 0.00 & 0.20 & 0.03 & 0.13 & 0.00 & 0.10 & 0.00 \\
    Steve-Eye-auto & 0.30 & 0.27 & 0.37 & 0.23 & 0.20 & 0.17 & 0.26 & 0.07 & 0.13 & 0.17 & 0.20 & 0.33 & 0.00 & 0.13 \\
    Steve-Eye & \textbf{0.40} & \textbf{0.30} & \textbf{0.43} & \textbf{0.53} & \textbf{0.33} & \textbf{0.37} & \textbf{0.43} & \textbf{0.30} & \textbf{0.43} & \textbf{0.47} & \textbf{0.47} & \textbf{0.40} & \textbf{0.13} & \textbf{0.23} \\
    
    \bottomrule
    \end{tabular}
    }
\end{subtable}
  
\vspace{0.1cm} 

\begin{subtable}{1\linewidth}
\centering
    \scalebox{0.8}{
    \begin{tabular}{l|cccccccccc}
    \toprule
    Model & \mcmilkbucket & \mcwool & \mcbeef & \mcmutton & \mcbed & \mcpainting & \mccarpet & \mcitemframe & \mccookedbeef & \mccookedmutton \\
    \midrule
    MineAgent & 0.46 & 0.50 & 0.33 & 0.35 & 0.0 & 0.0 & 0.06 & 0.0 & 0.0 & 0.0  \\
    gpt assistant & 0.57 & 0.76 & 0.43 & 0.30 & 0.00 & 0.00 & 0.37 & 0.00 & 0.03 & 0.00 \\
    Steve-Eye-auto  & 0.70 & 0.63 & 0.40 & 0.30 & 0.17 & 0 & 0.37 & 0.03 & 0.07 & 0.00 \\
    Steve-Eye  & \textbf{0.73} & 0.67 & \textbf{0.47} & 0.33 & \textbf{0.23} & \textbf{0.07} & \textbf{0.43} & \textbf{0.10} & \textbf{0.17} & \textbf{0.07} \\
    
    \bottomrule
    \end{tabular}
    }
\end{subtable}
\label{tab:skill_plan}
\end{table}

%% file: sections/5_conclusion.tex
\section{Conclusion}
In this paper, we explore enabling a large multimodal model to serve as a generative embodied agent in open worlds.
We achieve this goal by proposing Steve-Eye, which combines the text-only language model with a visual encoder, allowing for a multimodal I/O interface to interact with the environment. 
With the help of ChatGPT, we curate questions to generate 850K
instruction-following data to facilitate the agent's multimodal perception fuction, foundational knowledge mastery, as well as the capability of skill prediction and planning.
Experiments on three open-world benchmarks verify the advantages of our Steve-Eye over a wide range of perspectives.

%% file: sections/6_appendix.tex
\clearpage
\section{Appendix}
In this appendix, we offer a detailed introduction of the construction of our open-world instruction dataset, as outlined in Appendix~\ref{sec:sup_data}, including (1) multimodal perception instructions, (2) foundational knowledge instructions, (3) skill-related interaction instructions, and (4) template of instructional training data.
Furthermore, we delve into the skill planning benchmark and its associated task setups in Appendix~\ref{sec:sup_plan}.
In Appendix~\ref{sec:sup_mm_gen} we present qualitative cases that illustrate our model's ability to provide intuitive visual feedback and serve as an intelligent chatbot with a multimodal input-output interface.
Finally, we explore the potential applications of our model in diverse environments, such as Virtual Home~\citep{puig2018virtualhome}.

\subsection{Dataset}
\label{sec:sup_data}

\subsubsection{Multimodal Perception Instructions} 
\label{sec:sup_data_mm}
\begin{figure}[ht]
\centering
\includegraphics[width=1\textwidth]{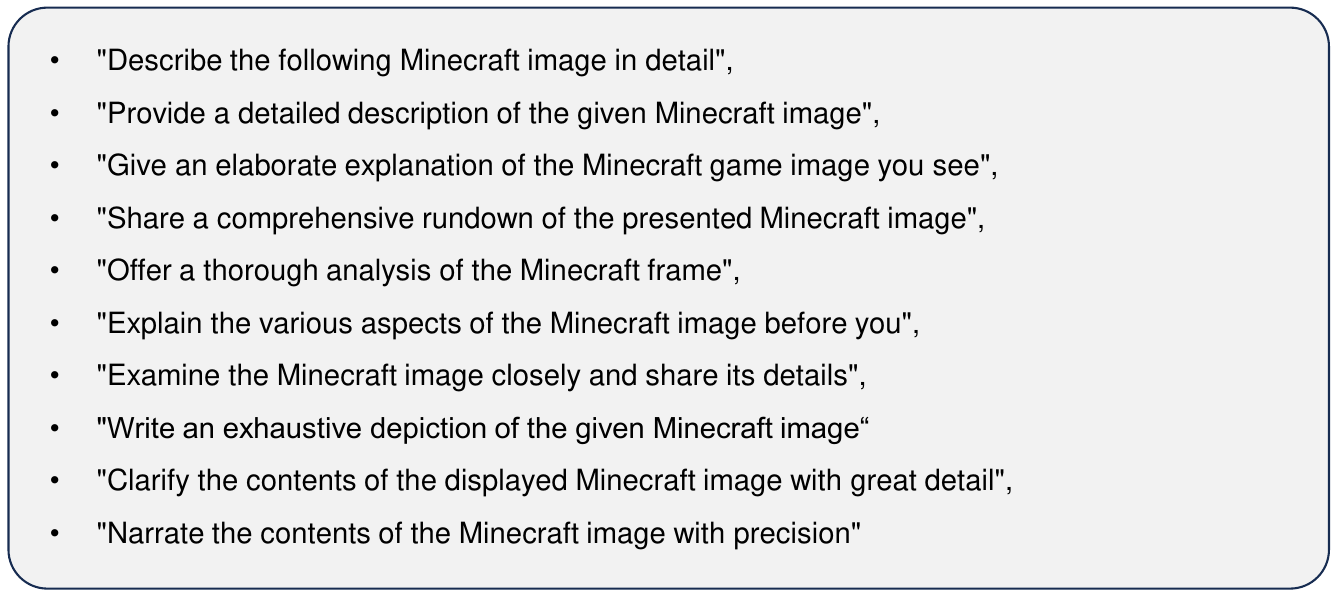}
\caption{
10 instruction examples for \textbf{multimodal perception instructions}. 
}
\label{fig:description_inst}
\end{figure}

This dataset contains 200K instructional pairs.
Figure~\ref{fig:description_inst} illustrates a partial listing of instructional questions employed for describing the content of the Minecraft snapshots.
These instructions convey similar meanings, albeit with slight variations in natural language.

\subsubsection{Foundational knowledge Instructions} 
\label{sec:sup_data_fk}
\begin{figure}[H]
\centering
\includegraphics[width=1\textwidth]{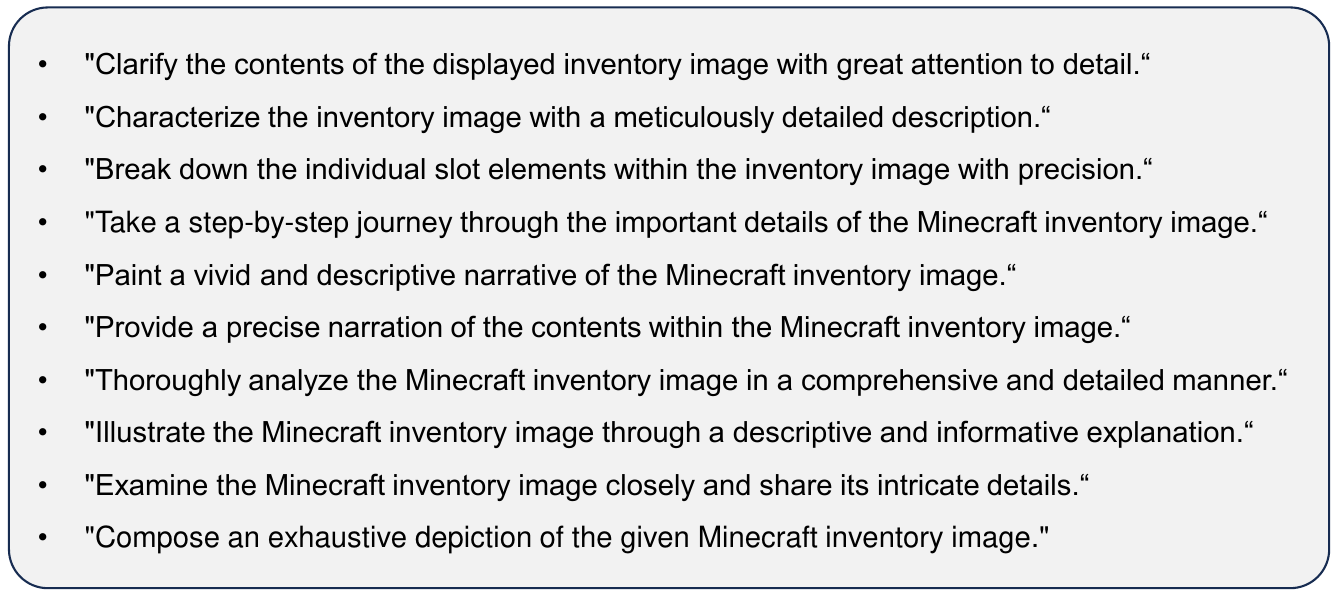}
\caption{
10 instruction examples of icon images for \textbf{foundational knowledge instructions}. 
}
\label{fig:icon_inst}
\end{figure}

\begin{figure}
\centering
\includegraphics[width=1\textwidth]{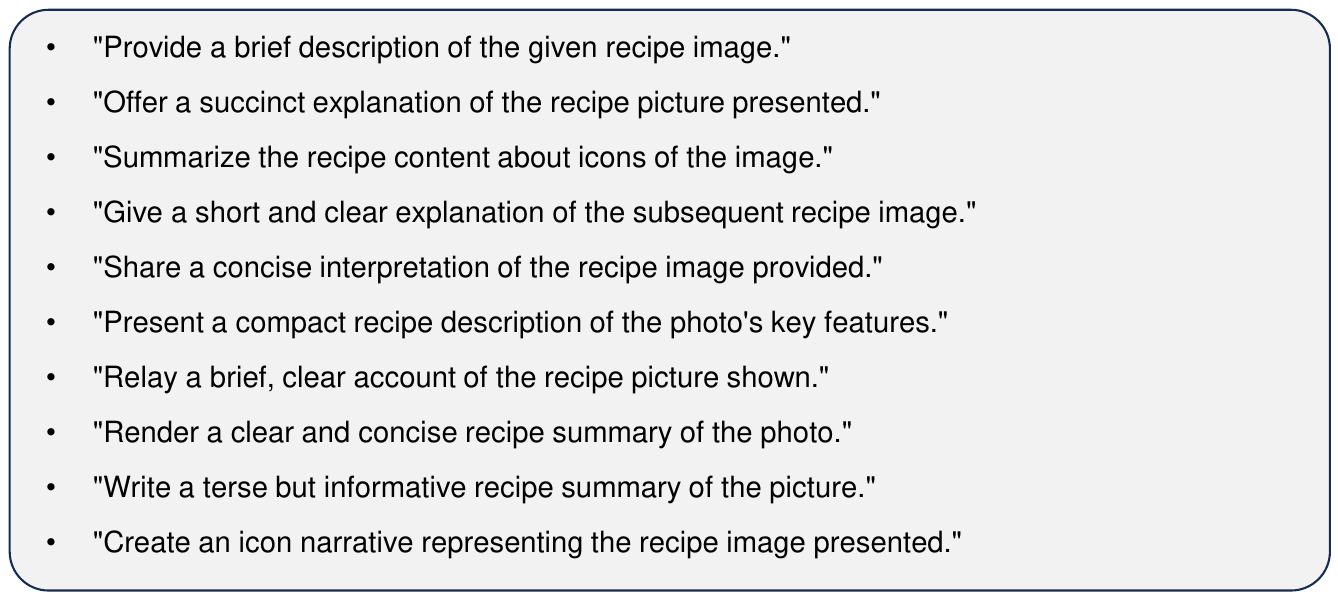}
\caption{
10 instruction examples of recipe image for \textbf{foundational knowledge instructions}. 
}
\label{fig:recipe_inst}
\end{figure}

The dataset comprises 250K training instances, which is organized into three distinct subsets: 200K icon image instructions, 10K recipe image instructions, and 40K Minecraft-Wiki corpus instructions.
For the icon images, we generate questions aimed at prompting the model to recognize and describe item icons within the inventory, as depicted in Figure~\ref{fig:icon_inst}.
Similarly, we curate instructional questions for recipe images as shown in Figure~\ref{fig:recipe_inst}, with the objective of extracting information on completing specific recipes.
In addition, we preprocess the raw Minecraft-Wiki HTML pages by removing irrelevant information (e.g., reference links) and unresolved data, transforming the raw corpus into a formatted, clean Markdown version.
Leveraging the capabilities of ChatGPT, we employ this powerful language model to generate 10 questions, each with its corresponding answer, for every page of the cleaned Wiki corpus. 
This process yields a collection of 40K single-round question-answer pairs, which can be utilized for instruction tuning.

\subsubsection{Skill-related Interaction Instructions}
\label{sec:sup_data_spp} 
\begin{figure}[H]
\centering
\includegraphics[width=1.\textwidth]{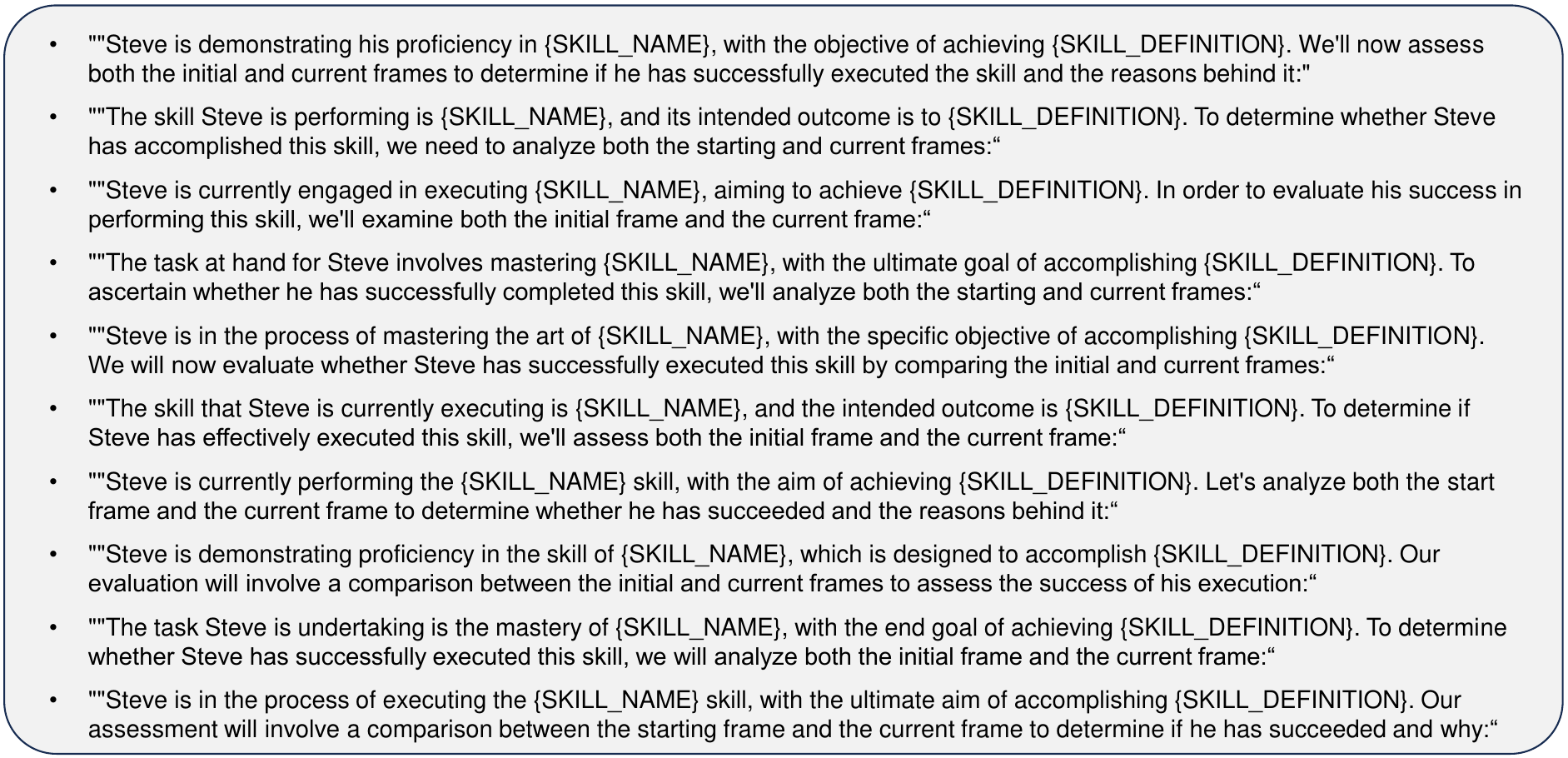}
\caption{
10 instruction examples for \textbf{skill prediction instructions}. 
}
\label{fig:skill_pred_inst}
\end{figure}

\begin{figure}[H]
\centering
\includegraphics[width=1\textwidth]{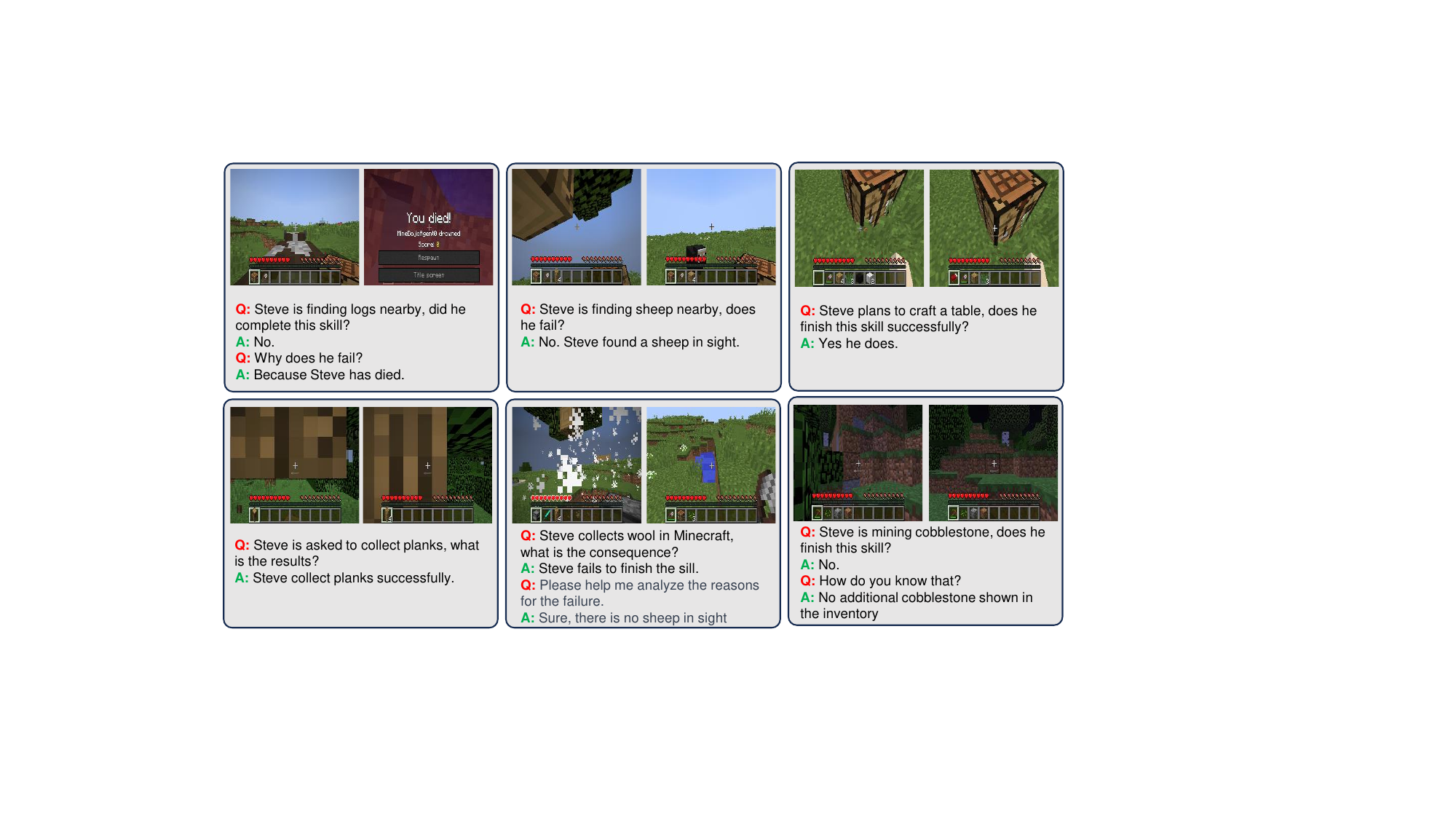}
\caption{
Illustrative examples of \textbf{skill prediction instruction} data with snapshot pairs. 
}
\label{fig:skill_pred_img}
\end{figure}

For skill prediction, we utilize the skill policies trained by~\cite{yuan2023plan4mc} to  create a dataset comprising 200K skill trajectories.
In each trajectory, we extract timestamps from the initial and t-th points to generate a snapshot pair, denoted as $\{\mathcal{I}_0, \mathcal{I}_t\}$.
We then construct questions aimed at determining whether the agent successfully executed the skill or, in the case of failure, identifying the underlying reasons for the unsuccessful attempt.
Illustrative examples of these skill prediction questions are provided in Figure~\ref{fig:skill_pred_inst}.
We also  provide examples with snapshot pairs in Figure~\ref{fig:skill_pred_img}.

\begin{figure}[H]
\centering
\includegraphics[width=0.7\textwidth]{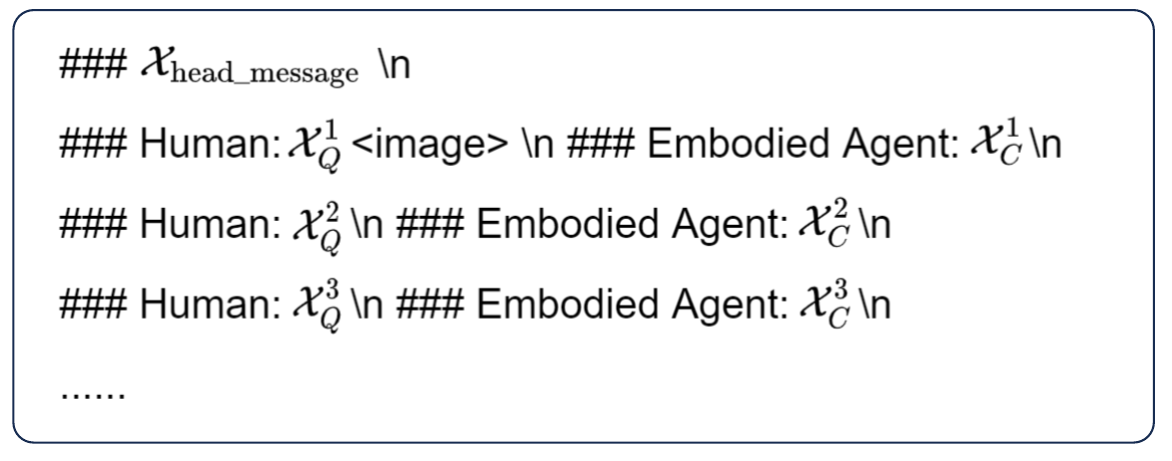}
\caption{The unified template to generate input sequence for instructional tuning.}
\label{fig:uni_prompt}
\end{figure}

\noindent\textbf{Template of Instructional Training Data. }
\label{sec:sup_data_train}
Similar to \citet{liu2023visual}, we formulate each instructional instance as a multi-round conversation as shown in Figure~\ref{fig:uni_prompt}, where $\mathcal{X}_{\text{head message}}$ is a sentence to describe this assistant (e.g., 	`` You are in a chat between a curious human and an artificial intelligence assistant. You should serve as an assistant to give helpful, detailed, and polite answers to the human’s questions.'').
The number of rounds relies on the input instruction content.
And the input images (denoted as $<\text{image}>$) will only be fed in the first round, while $\mathcal{X}_C$ may contain visual outputs with two additional tokens $<\text{vis}>$ and $</\text{vis}>$.

\subsection{Skill Planning Benchmark}
\label{sec:sup_plan}
To clarify this benchmark, we begin by offering comprehensive task setup details in Table~\ref{tab:task_setup}.
During the evaluation phase, we relocate the agent to a random location at the initiation of every episode, with distances of up to 500 units, ensuring that the agent spawns in an unfamiliar environment.
Furthermore, for tasks that involve interacting with mobs, we enforce a maximum spawning distance of 30 units for cows and sheep.
Our approach to complete tasks is rooted in a hierarchical framework. 
Specifically, our model exclusively generates high-level skill plans, delegating the actual skill execution to pre-trained basic skill policies as introduced by \citet{yuan2023plan4mc}.
Notably, we introduce a self-driven variant named 'Steve-Eye-auto,' which serves not only as a planner but also replaces the Minecraft rules to verify the successful execution of skills.

\input{tables/task-desc}

\begin{figure}[t]
\centering
\begin{subfigure}{1\textwidth}
\includegraphics[width=.9\textwidth]{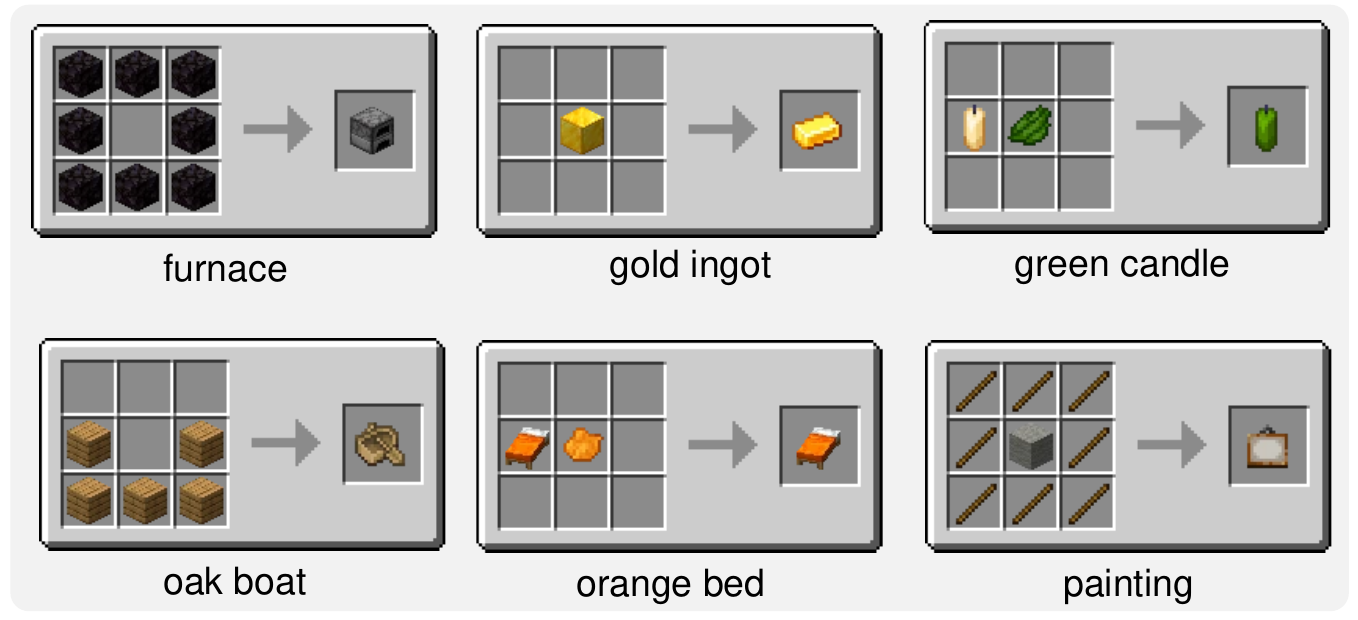}
\caption{
Qualitative examples of recipe image generation.
} 
\label{fig:recipe_image_gen}
\end{subfigure}
\vspace{0.1cm} 

\begin{subfigure}{1\textwidth}
\includegraphics[width=.9\textwidth]{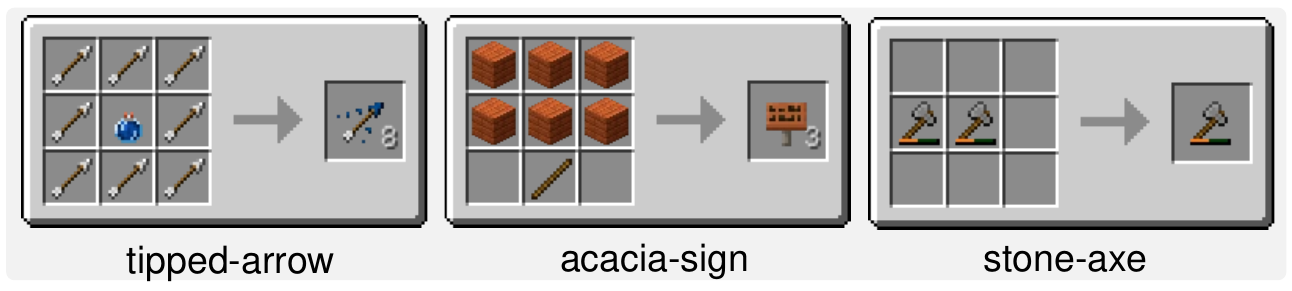}
\caption{
Illustrative examples of recipes that our model struggles to predict accurately. We attribute this failure to the complexities arising from fine-grained or semantically overlapping image information.
} 
\label{fig:recipe_image_gen_2}
\end{subfigure}

\end{figure}

\begin{figure}[H]
\centering
\includegraphics[width=1\textwidth]{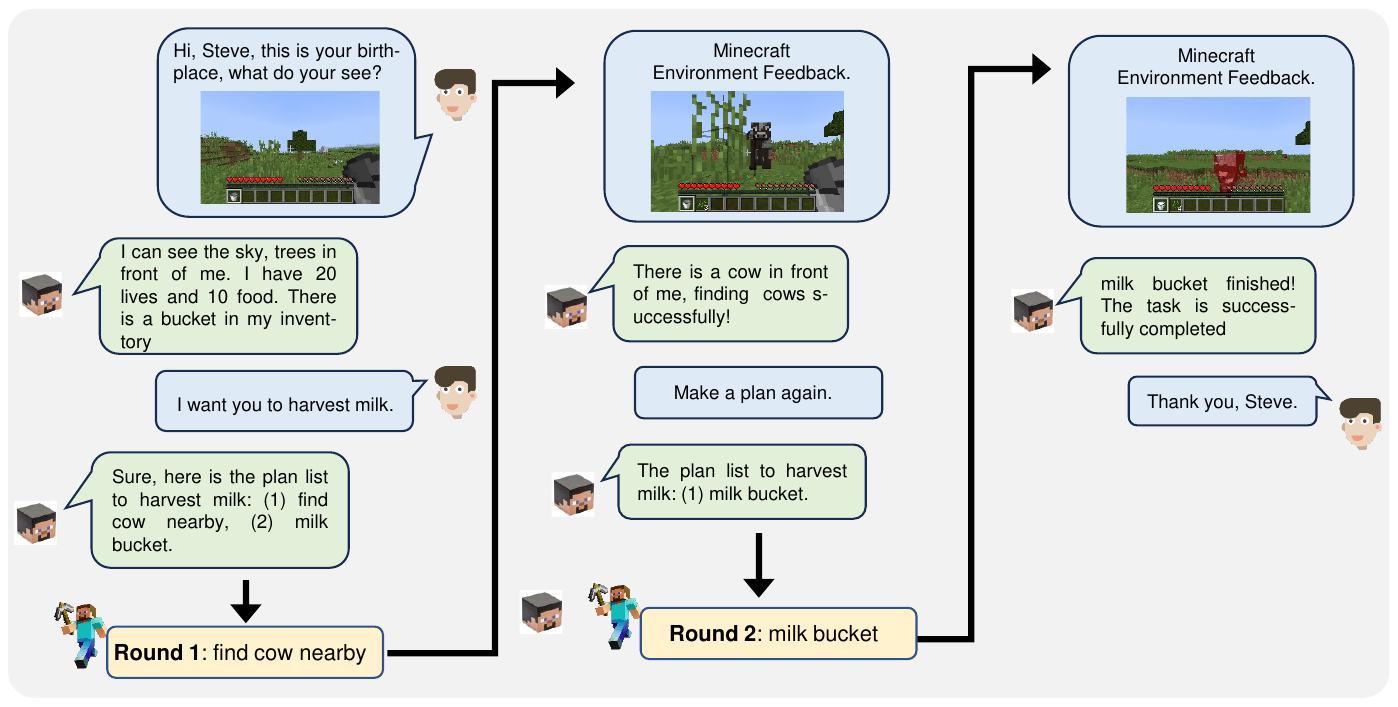}
\caption{An overview of the process through which the chatbot receives and fulfills a task command.}
\label{fig:chatbot}
\end{figure}

\subsection{Qualitative results of Multimodal Generation}
\label{sec:sup_mm_gen}

\subsubsection{Recipe Image Generation}
Figure~\ref{fig:recipe_image_gen} showcases qualitative examples of our evaluation on the FK-QA IMG dataset.
Utilizing a visual tokenizer like VG-GAN, our model demonstrates the ability to engage in visual generation, enabling it to provide visual feedback based on its comprehension of textual input.
However, as shown in Figure~\ref{fig:recipe_image_gen_2}, our model encounters difficulties when generating image content characterized by fine-grained or semantically overlapping elements.
These challenges warrant further exploration in our future work.

\subsubsection{Multimodal ChatBot}
In Figure~\ref{fig:chatbot}, We present an overview of Steve-Eye functioning as a chatbot to receive task commands and execute them.

\begin{figure}[t]
\includegraphics[width=1.\textwidth]{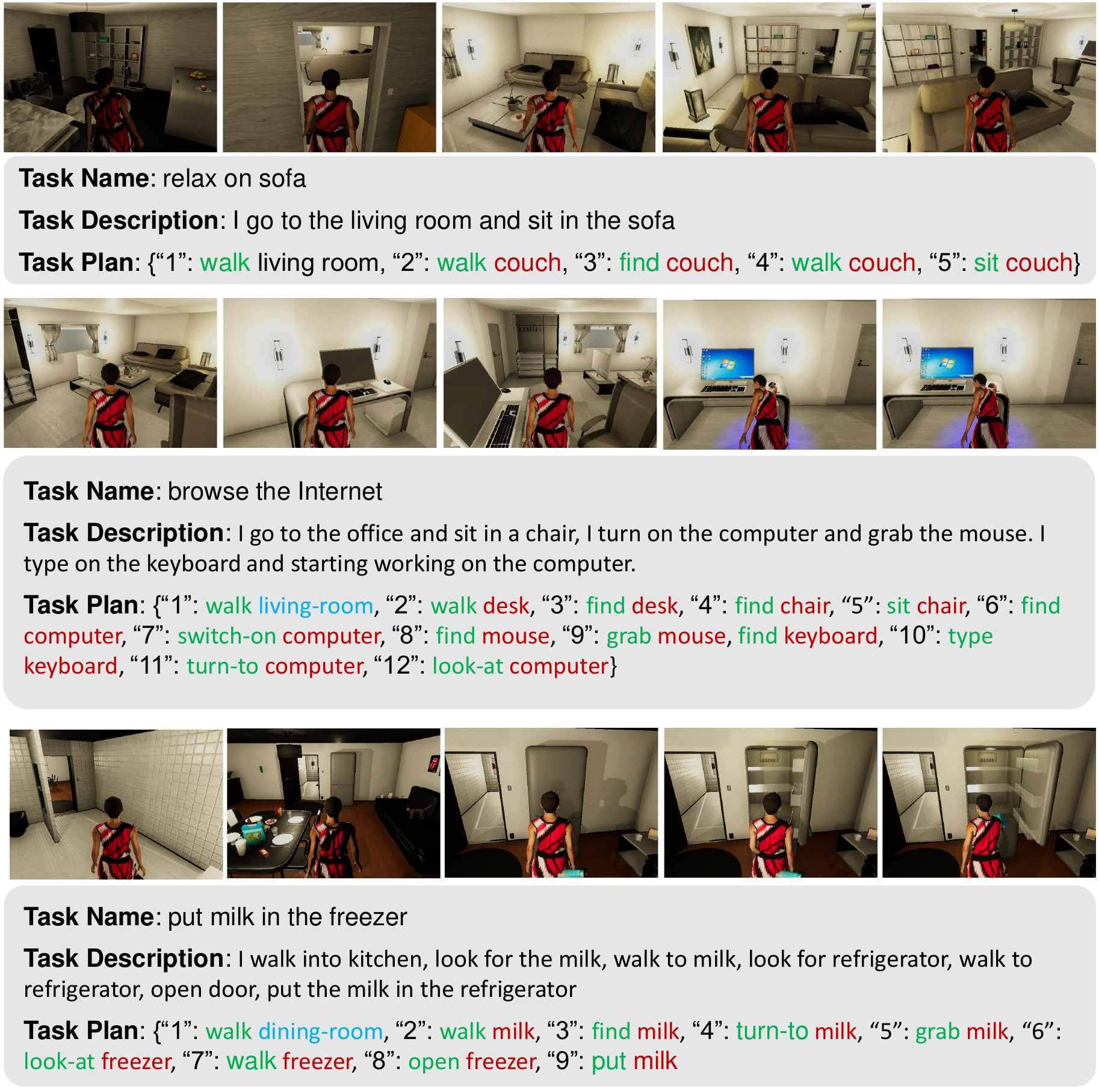}
\caption{
Task examples from the extended Virtual-Home benchmark, where elements in \textcolor{green!70!black}{green}, \textcolor{cyan}{cyan} and \textcolor{red!75!black}{red} represent action, room, and object categories, respectively.
Our benchmark includes a diverse range of tasks that simulate interactions between individuals and their room environments.  
It contains over 50 distinct room setups, involving 20 unique actions, and 100 objects. 
Each room presents a selection of more than 200 distinct tasks.} 
\label{fig:virtual_home}
\end{figure}

\subsection{Discussion of Open-World Exploration}
\label{sec:sup_env}

In this paper, we have selected Minecraft as our open-world platform.
Nevertheless, it is evident that Steve-Eye can be applied to other open-world environments, such as Virtual Home~\citep{puig2018virtualhome} and AI2THOR~\citep{kolve2017ai2}, with minimal manual effort using the same methodology in this paper.
These alternative benchmarks, when compared to Minecraft, exhibit a closer alignment with the real world. 
To some extent, this choice holds greater significance since our ultimate objective is to deploy the agent in the real world.
To achieve this goal, we expand the Virtual Home benchmark by introducing a more extensive range of environments (50+ rooms), human-interaction tasks (200+ for each room), as well as diverse categories of actions (20+) and objects (100+), as illustrated in Figure~\ref{fig:virtual_home}.
The corresponding validation and further exploration of open-ended embodied agents in a real-world context will be the focus of our future work.

%% file: tables/task-desc.tex
\renewcommand{\arraystretch}{1.4}
\begin{table}[h]
	\centering
	\caption{The setups of 24 tasks used in our skill planning evaluation, where ``Initial Items'' refers to the tools provided in the agent's inventory at the beginning of each episode, and ``Max Steps'' represents the maximum episode duration. Any episode exceeding this limit is classified as a task failure. The tasks are originally developed by~\citet{yuan2023plan4mc}.}
	\centering
	\begin{subtable}{1\linewidth}
		\centering
		\caption{7 tasks  involving the process of ``cutting trees to craft primary items''.}
		\scalebox{0.8}{
			\begin{tabular}{l|ccccccc}
				\toprule
				Task Icon & \mcstick & \mccraftingtable & \mcbowl & \mcchest & \mctrapdoor & \mcsign 
				& \mcwoodenpickaxe  \\
				\midrule\hline
				Task Name & stick & crafting$\_$table$\_$nearby & bowl & chest & trap$\_$door & sign & wooden$\_$pickaxe \\
				Initial Items & - & - & - & - & - & - & - \\
				Max Steps & 3000  & 3000  & 3000  & 3000  & 3000  & 3000  & 3000 \\
				\bottomrule
			\end{tabular}
		}
	\end{subtable}
	
	\vspace{0.3cm} 
	
	\begin{subtable}{1\linewidth}
		\centering
		\caption{7 tasks  involving the process of ``mining cobblestones to craft advanced items''.}
		\scalebox{0.8}{
			\begin{tabular}{l|ccccccc}
				\toprule
				Task Icon & \mcfurnace & \mcstonestairs & \mcstoneslab & \mccobblestonewall & \mclever & \mctorch & \mcstonepickaxe \\
				\midrule
				
				\hline
				Task Name & furname$\_$nearby & stone$\_$stairs & stone$\_$slab & cobblestone$\_$wall & lever & torch & stone$\_$pickaxe \\
				Initial Items & \mclog*10 & \mclog*10 & \mclog*10 & \mclog*10 & \mcwoodenpickaxe & \mclog*10  &  \mcwoodenpickaxe  \\
				Max Steps & 5000 & 5000 & 3000 & 5000 & 5000 & 5000 & 10000 \\
				\bottomrule
			\end{tabular}
		}
	\end{subtable}
	
	\vspace{0.3cm} 
	
	\begin{subtable}{1\linewidth}
		\centering
		\caption{10 tasks  involving the process of ``interacting with mobs to harvest food and materials''.}
		\scalebox{0.8}{
			\begin{tabular}{l|p{1cm}ccccccp{1cm}p{1cm}p{1cm}}
				\toprule
				Task Icon & \mcmilkbucket & \mcwool & \mcbeef & \mcmutton & \mcbed & \mcpainting & \mccarpet & \mcitemframe & \mccookedbeef & \mccookedmutton \\
				\midrule\hline
				Task Name & milk$\_$ bucket & wool & beef & mutton & bed & painting & carpet & item$\_$ frame & cooked$\_$ beef & cooked$\_$ button \\
				Initial Items &  \mccraftingtable, \mcironingot*3 &  \mccraftingtable, \mcironingot*2 & \mcdiamondsword & \mcdiamondsword &  \mccraftingtable, \mcshears &  \mccraftingtable, \mcshears & \mcshears & \mccraftingtable, \mcdiamondsword & \mccraftingtable, \mcdiamondsword  & \mcfurnace, \mcdiamondsword \\
				Max Steps & 3000 & 3000 & 3000 & 3000 & 10000 & 10000 & 3000 & 10000 & 10000 & 10000 \\
				\bottomrule
			\end{tabular}
		}
	\end{subtable}
\label{tab:task_setup}
\end{table}
\renewcommand{\arraystretch}{1}

\begin{table}[htb]
\caption{
The setups of 10 long-horizon iron-based tasks, where ``Initial Items'' are provided in the agent's inventory at task beginning, and ``Max Steps'' refers to maximum environmental steps.}
\centering
\scalebox{0.9}{
\begin{tabular}{ccccc}
\toprule
Task icon & Task description & Initial tools & Max steps \\
\midrule
\mcironingot & craft iron ingot & \mcstonepickaxe *5, \mcdirt *64  & 8000 \\
\mcshears & craft shears & \mcstonepickaxe *5, \mcdirt *64  & 10000 \\
\mcbucket & craft bucket & \mcstonepickaxe *5, \mcdirt *64 & 12000 \\
\mcironpickaxe & craft iron pickaxe & \mcstonepickaxe *5, \mcdirt *64 & 12000 \\
\mcironaxe & craft iron axe & \mcstonepickaxe *5, \mcdirt *64  & 12000 \\
\mcironsword & craft iron sword & \mcstonepickaxe *5, \mcdirt *64  & 10000 \\
\mcironshovel & craft iron shovel & \mcstonepickaxe *5, \mcdirt *64 & 8000 \\
\mctripwirehook & craft tripwire hook & \mcstonepickaxe *5, \mcdirt *64 & 8000 \\
\mcheavypressureplate & craft heavy weighted pressure plate & \mcstonepickaxe *5, \mcdirt *64 t & 10000 \\
\mcirontrapdoor & craft iron trapdoor & \mcstonepickaxe *5, \mcdirt *64  & 12000 \\
\bottomrule
\end{tabular}}
\end{table}

%% file: iclr2024_conference.bbl
\begin{thebibliography}{42}
\providecommand{\natexlab}[1]{#1}
\providecommand{\url}[1]{\texttt{#1}}
\expandafter\ifx\csname urlstyle\endcsname\relax
  \providecommand{\doi}[1]{doi: #1}\else
  \providecommand{\doi}{doi: \begingroup \urlstyle{rm}\Url}\fi

\bibitem[Ahn et~al.(2022)Ahn, Brohan, Brown, Chebotar, Cortes, David, Finn, Fu,
  Gopalakrishnan, Hausman, et~al.]{ahn2022can}
Michael Ahn, Anthony Brohan, Noah Brown, Yevgen Chebotar, Omar Cortes, Byron
  David, Chelsea Finn, Chuyuan Fu, Keerthana Gopalakrishnan, Karol Hausman,
  et~al.
\newblock Do as i can, not as i say: Grounding language in robotic affordances.
\newblock \emph{arXiv preprint arXiv:2204.01691}, 2022.

\bibitem[Alayrac et~al.(2022)Alayrac, Donahue, Luc, Miech, Barr, Hasson, Lenc,
  Mensch, Millican, Reynolds, et~al.]{alayrac2022flamingo}
Jean-Baptiste Alayrac, Jeff Donahue, Pauline Luc, Antoine Miech, Iain Barr,
  Yana Hasson, Karel Lenc, Arthur Mensch, Katherine Millican, Malcolm Reynolds,
  et~al.
\newblock Flamingo: a visual language model for few-shot learning.
\newblock \emph{Advances in Neural Information Processing Systems},
  35:\penalty0 23716--23736, 2022.

\bibitem[Awadalla et~al.(2023)Awadalla, Gao, Gardner, Hessel, Hanafy, Zhu,
  Marathe, Bitton, Gadre, Sagawa, et~al.]{awadalla2023openflamingo}
Anas Awadalla, Irena Gao, Josh Gardner, Jack Hessel, Yusuf Hanafy, Wanrong Zhu,
  Kalyani Marathe, Yonatan Bitton, Samir Gadre, Shiori Sagawa, et~al.
\newblock Openflamingo: An open-source framework for training large
  autoregressive vision-language models.
\newblock \emph{arXiv preprint arXiv:2308.01390}, 2023.

\bibitem[Bommasani et~al.(2021)Bommasani, Hudson, Adeli, Altman, Arora, von
  Arx, Bernstein, Bohg, Bosselut, Brunskill,
  et~al.]{bommasani2021opportunities}
Rishi Bommasani, Drew~A Hudson, Ehsan Adeli, Russ Altman, Simran Arora, Sydney
  von Arx, Michael~S Bernstein, Jeannette Bohg, Antoine Bosselut, Emma
  Brunskill, et~al.
\newblock On the opportunities and risks of foundation models.
\newblock \emph{arXiv preprint arXiv:2108.07258}, 2021.

\bibitem[Brown et~al.(2020)Brown, Mann, Ryder, Subbiah, Kaplan, Dhariwal,
  Neelakantan, Shyam, Sastry, Askell, et~al.]{brown2020language}
Tom Brown, Benjamin Mann, Nick Ryder, Melanie Subbiah, Jared~D Kaplan, Prafulla
  Dhariwal, Arvind Neelakantan, Pranav Shyam, Girish Sastry, Amanda Askell,
  et~al.
\newblock Language models are few-shot learners.
\newblock \emph{Advances in neural information processing systems},
  33:\penalty0 1877--1901, 2020.

\bibitem[Chowdhery et~al.(2022)Chowdhery, Narang, Devlin, Bosma, Mishra,
  Roberts, Barham, Chung, Sutton, Gehrmann, et~al.]{chowdhery2022palm}
Aakanksha Chowdhery, Sharan Narang, Jacob Devlin, Maarten Bosma, Gaurav Mishra,
  Adam Roberts, Paul Barham, Hyung~Won Chung, Charles Sutton, Sebastian
  Gehrmann, et~al.
\newblock Palm: Scaling language modeling with pathways.
\newblock \emph{arXiv preprint arXiv:2204.02311}, 2022.

\bibitem[Driess et~al.(2023)Driess, Xia, Sajjadi, Lynch, Chowdhery, Ichter,
  Wahid, Tompson, Vuong, Yu, et~al.]{driess2023palm}
Danny Driess, Fei Xia, Mehdi~SM Sajjadi, Corey Lynch, Aakanksha Chowdhery,
  Brian Ichter, Ayzaan Wahid, Jonathan Tompson, Quan Vuong, Tianhe Yu, et~al.
\newblock Palm-e: An embodied multimodal language model.
\newblock \emph{arXiv preprint arXiv:2303.03378}, 2023.

\bibitem[Esser et~al.(2021)Esser, Rombach, and Ommer]{esser2021taming}
Patrick Esser, Robin Rombach, and Bjorn Ommer.
\newblock Taming transformers for high-resolution image synthesis.
\newblock In \emph{Proceedings of the IEEE/CVF conference on computer vision
  and pattern recognition}, pp.\  12873--12883, 2021.

\bibitem[Fallman(2003)]{fallman2003design}
Daniel Fallman.
\newblock Design-oriented human-computer interaction.
\newblock In \emph{Proceedings of the SIGCHI conference on Human factors in
  computing systems}, pp.\  225--232, 2003.

\bibitem[Fan et~al.(2022)Fan, Wang, Jiang, Mandlekar, Yang, Zhu, Tang, Huang,
  Zhu, and Anandkumar]{fan2022minedojo}
Linxi Fan, Guanzhi Wang, Yunfan Jiang, Ajay Mandlekar, Yuncong Yang, Haoyi Zhu,
  Andrew Tang, De-An Huang, Yuke Zhu, and Anima Anandkumar.
\newblock Minedojo: Building open-ended embodied agents with internet-scale
  knowledge.
\newblock \emph{Advances in Neural Information Processing Systems},
  35:\penalty0 18343--18362, 2022.

\bibitem[Fandom(2023)]{mcwiki}
Fandom.
\newblock Minecraft wiki.
\newblock \emph{https://minecraft.fandom.com/wiki}, 2023.

\bibitem[Gupta \& Kembhavi(2023)Gupta and Kembhavi]{gupta2023visual}
Tanmay Gupta and Aniruddha Kembhavi.
\newblock Visual programming: Compositional visual reasoning without training.
\newblock In \emph{Proceedings of the IEEE/CVF Conference on Computer Vision
  and Pattern Recognition}, pp.\  14953--14962, 2023.

\bibitem[Guss et~al.(2019)Guss, Houghton, Topin, Wang, Codel, Veloso, and
  Salakhutdinov]{guss2019minerl}
William~H Guss, Brandon Houghton, Nicholay Topin, Phillip Wang, Cayden Codel,
  Manuela Veloso, and Ruslan Salakhutdinov.
\newblock Minerl: A large-scale dataset of minecraft demonstrations.
\newblock \emph{arXiv preprint arXiv:1907.13440}, 2019.

\bibitem[Hu et~al.(2021)Hu, Shen, Wallis, Allen-Zhu, Li, Wang, Wang, and
  Chen]{hu2021lora}
Edward~J Hu, Yelong Shen, Phillip Wallis, Zeyuan Allen-Zhu, Yuanzhi Li, Shean
  Wang, Lu~Wang, and Weizhu Chen.
\newblock Lora: Low-rank adaptation of large language models.
\newblock \emph{arXiv preprint arXiv:2106.09685}, 2021.

\bibitem[Huang et~al.(2023{\natexlab{a}})Huang, Li, Yang, Shi, Chang, Ye, Wu,
  Hong, Huang, Liu, et~al.]{huang2023audiogpt}
Rongjie Huang, Mingze Li, Dongchao Yang, Jiatong Shi, Xuankai Chang, Zhenhui
  Ye, Yuning Wu, Zhiqing Hong, Jiawei Huang, Jinglin Liu, et~al.
\newblock Audiogpt: Understanding and generating speech, music, sound, and
  talking head.
\newblock \emph{arXiv preprint arXiv:2304.12995}, 2023{\natexlab{a}}.

\bibitem[Huang et~al.(2023{\natexlab{b}})Huang, Dong, Wang, Hao, Singhal, Ma,
  Lv, Cui, Mohammed, Liu, et~al.]{huang2023language}
Shaohan Huang, Li~Dong, Wenhui Wang, Yaru Hao, Saksham Singhal, Shuming Ma,
  Tengchao Lv, Lei Cui, Owais~Khan Mohammed, Qiang Liu, et~al.
\newblock Language is not all you need: Aligning perception with language
  models.
\newblock \emph{arXiv preprint arXiv:2302.14045}, 2023{\natexlab{b}}.

\bibitem[Huang et~al.(2022{\natexlab{a}})Huang, Abbeel, Pathak, and
  Mordatch]{huang2022language}
Wenlong Huang, Pieter Abbeel, Deepak Pathak, and Igor Mordatch.
\newblock Language models as zero-shot planners: Extracting actionable
  knowledge for embodied agents.
\newblock In \emph{International Conference on Machine Learning}, pp.\
  9118--9147. PMLR, 2022{\natexlab{a}}.

\bibitem[Huang et~al.(2022{\natexlab{b}})Huang, Xia, Xiao, Chan, Liang,
  Florence, Zeng, Tompson, Mordatch, Chebotar, et~al.]{huang2022inner}
Wenlong Huang, Fei Xia, Ted Xiao, Harris Chan, Jacky Liang, Pete Florence, Andy
  Zeng, Jonathan Tompson, Igor Mordatch, Yevgen Chebotar, et~al.
\newblock Inner monologue: Embodied reasoning through planning with language
  models.
\newblock \emph{arXiv preprint arXiv:2207.05608}, 2022{\natexlab{b}}.

\bibitem[Kolve et~al.(2017)Kolve, Mottaghi, Han, VanderBilt, Weihs, Herrasti,
  Deitke, Ehsani, Gordon, Zhu, et~al.]{kolve2017ai2}
Eric Kolve, Roozbeh Mottaghi, Winson Han, Eli VanderBilt, Luca Weihs, Alvaro
  Herrasti, Matt Deitke, Kiana Ehsani, Daniel Gordon, Yuke Zhu, et~al.
\newblock Ai2-thor: An interactive 3d environment for visual ai.
\newblock \emph{arXiv preprint arXiv:1712.05474}, 2017.

\bibitem[Li et~al.(2023)Li, Li, Savarese, and Hoi]{li2023blip}
Junnan Li, Dongxu Li, Silvio Savarese, and Steven Hoi.
\newblock Blip-2: Bootstrapping language-image pre-training with frozen image
  encoders and large language models.
\newblock \emph{arXiv preprint arXiv:2301.12597}, 2023.

\bibitem[Li et~al.(2022)Li, Puig, Paxton, Du, Wang, Fan, Chen, Huang,
  Aky{\"u}rek, Anandkumar, et~al.]{li2022pre}
Shuang Li, Xavier Puig, Chris Paxton, Yilun Du, Clinton Wang, Linxi Fan, Tao
  Chen, De-An Huang, Ekin Aky{\"u}rek, Anima Anandkumar, et~al.
\newblock Pre-trained language models for interactive decision-making.
\newblock \emph{Advances in Neural Information Processing Systems},
  35:\penalty0 31199--31212, 2022.

\bibitem[Liu et~al.(2023)Liu, Li, Wu, and Lee]{liu2023visual}
Haotian Liu, Chunyuan Li, Qingyang Wu, and Yong~Jae Lee.
\newblock Visual instruction tuning.
\newblock \emph{arXiv preprint arXiv:2304.08485}, 2023.

\bibitem[OpenAI(2022)]{chatgpt}
OpenAI.
\newblock Chatgpt.
\newblock \emph{https://openai.com/blog/chatgpt}, 2022.

\bibitem[OpenAI(2023)]{gpt4}
OpenAI.
\newblock Gpt-4 technical report.
\newblock \emph{arXiv:2303.08774}, 2023.

\bibitem[Park et~al.(2023)Park, O'Brien, Cai, Morris, Liang, and
  Bernstein]{park2023generative}
Joon~Sung Park, Joseph~C O'Brien, Carrie~J Cai, Meredith~Ringel Morris, Percy
  Liang, and Michael~S Bernstein.
\newblock Generative agents: Interactive simulacra of human behavior.
\newblock \emph{arXiv preprint arXiv:2304.03442}, 2023.

\bibitem[Patil et~al.(2023)Patil, Zhang, Wang, and Gonzalez]{patil2023gorilla}
Shishir~G Patil, Tianjun Zhang, Xin Wang, and Joseph~E Gonzalez.
\newblock Gorilla: Large language model connected with massive apis.
\newblock \emph{arXiv preprint arXiv:2305.15334}, 2023.

\bibitem[Peng et~al.(2023)Peng, Wang, Dong, Hao, Huang, Ma, and
  Wei]{peng2023kosmos}
Zhiliang Peng, Wenhui Wang, Li~Dong, Yaru Hao, Shaohan Huang, Shuming Ma, and
  Furu Wei.
\newblock Kosmos-2: Grounding multimodal large language models to the world.
\newblock \emph{arXiv preprint arXiv:2306.14824}, 2023.

\bibitem[Preece et~al.(1994)Preece, Rogers, Sharp, Benyon, Holland, and
  Carey]{preece1994human}
Jenny Preece, Yvonne Rogers, Helen Sharp, David Benyon, Simon Holland, and Tom
  Carey.
\newblock \emph{Human-computer interaction}.
\newblock Addison-Wesley Longman Ltd., 1994.

\bibitem[Puig et~al.(2018)Puig, Ra, Boben, Li, Wang, Fidler, and
  Torralba]{puig2018virtualhome}
Xavier Puig, Kevin Ra, Marko Boben, Jiaman Li, Tingwu Wang, Sanja Fidler, and
  Antonio Torralba.
\newblock Virtualhome: Simulating household activities via programs.
\newblock In \emph{Proceedings of the IEEE Conference on Computer Vision and
  Pattern Recognition}, pp.\  8494--8502, 2018.

\bibitem[Radford et~al.(2021)Radford, Kim, Hallacy, Ramesh, Goh, Agarwal,
  Sastry, Askell, Mishkin, Clark, et~al.]{radford2021learning}
Alec Radford, Jong~Wook Kim, Chris Hallacy, Aditya Ramesh, Gabriel Goh,
  Sandhini Agarwal, Girish Sastry, Amanda Askell, Pamela Mishkin, Jack Clark,
  et~al.
\newblock Learning transferable visual models from natural language
  supervision.
\newblock In \emph{International conference on machine learning}, pp.\
  8748--8763. PMLR, 2021.

\bibitem[Raffel et~al.(2020)Raffel, Shazeer, Roberts, Lee, Narang, Matena,
  Zhou, Li, and Liu]{raffel2020exploring}
Colin Raffel, Noam Shazeer, Adam Roberts, Katherine Lee, Sharan Narang, Michael
  Matena, Yanqi Zhou, Wei Li, and Peter~J Liu.
\newblock Exploring the limits of transfer learning with a unified text-to-text
  transformer.
\newblock \emph{The Journal of Machine Learning Research}, 21\penalty0
  (1):\penalty0 5485--5551, 2020.

\bibitem[Savva et~al.(2019)Savva, Kadian, Maksymets, Zhao, Wijmans, Jain,
  Straub, Liu, Koltun, Malik, et~al.]{savva2019habitat}
Manolis Savva, Abhishek Kadian, Oleksandr Maksymets, Yili Zhao, Erik Wijmans,
  Bhavana Jain, Julian Straub, Jia Liu, Vladlen Koltun, Jitendra Malik, et~al.
\newblock Habitat: A platform for embodied ai research.
\newblock In \emph{Proceedings of the IEEE/CVF international conference on
  computer vision}, pp.\  9339--9347, 2019.

\bibitem[Sur{\'\i}s et~al.(2023)Sur{\'\i}s, Menon, and
  Vondrick]{suris2023vipergpt}
D{\'\i}dac Sur{\'\i}s, Sachit Menon, and Carl Vondrick.
\newblock Vipergpt: Visual inference via python execution for reasoning.
\newblock \emph{arXiv preprint arXiv:2303.08128}, 2023.

\bibitem[Team(2022)]{vicuna}
Vicuna Team.
\newblock Vicuna: An open-source chatbot impressing gpt-4 with 90
  quality.
\newblock \emph{https://vicuna.lmsys.org/}, 2022.

\bibitem[Touvron et~al.(2023{\natexlab{a}})Touvron, Lavril, Izacard, Martinet,
  Lachaux, Lacroix, Rozi{\`e}re, Goyal, Hambro, Azhar,
  et~al.]{touvron2023llama}
Hugo Touvron, Thibaut Lavril, Gautier Izacard, Xavier Martinet, Marie-Anne
  Lachaux, Timoth{\'e}e Lacroix, Baptiste Rozi{\`e}re, Naman Goyal, Eric
  Hambro, Faisal Azhar, et~al.
\newblock Llama: Open and efficient foundation language models.
\newblock \emph{arXiv preprint arXiv:2302.13971}, 2023{\natexlab{a}}.

\bibitem[Touvron et~al.(2023{\natexlab{b}})Touvron, Martin, Stone, Albert,
  Almahairi, Babaei, Bashlykov, Batra, Bhargava, Bhosale,
  et~al.]{touvron2023llama2}
Hugo Touvron, Louis Martin, Kevin Stone, Peter Albert, Amjad Almahairi, Yasmine
  Babaei, Nikolay Bashlykov, Soumya Batra, Prajjwal Bhargava, Shruti Bhosale,
  et~al.
\newblock Llama 2: Open foundation and fine-tuned chat models.
\newblock \emph{arXiv preprint arXiv:2307.09288}, 2023{\natexlab{b}}.

\bibitem[Wang et~al.(2023{\natexlab{a}})Wang, Xie, Jiang, Mandlekar, Xiao, Zhu,
  Fan, and Anandkumar]{wang2023voyager}
Guanzhi Wang, Yuqi Xie, Yunfan Jiang, Ajay Mandlekar, Chaowei Xiao, Yuke Zhu,
  Linxi Fan, and Anima Anandkumar.
\newblock Voyager: An open-ended embodied agent with large language models.
\newblock \emph{arXiv preprint arXiv:2305.16291}, 2023{\natexlab{a}}.

\bibitem[Wang et~al.(2023{\natexlab{b}})Wang, Cai, Liu, Ma, and
  Liang]{wang2023describe}
Zihao Wang, Shaofei Cai, Anji Liu, Xiaojian Ma, and Yitao Liang.
\newblock Describe, explain, plan and select: Interactive planning with large
  language models enables open-world multi-task agents.
\newblock \emph{arXiv preprint arXiv:2302.01560}, 2023{\natexlab{b}}.

\bibitem[Yuan et~al.(2023)Yuan, Zhang, Wang, Xie, Cai, Dong, and
  Lu]{yuan2023plan4mc}
Haoqi Yuan, Chi Zhang, Hongcheng Wang, Feiyang Xie, Penglin Cai, Hao Dong, and
  Zongqing Lu.
\newblock Plan4mc: Skill reinforcement learning and planning for open-world
  minecraft tasks.
\newblock \emph{arXiv preprint arXiv:2303.16563}, 2023.

\bibitem[Zhang et~al.(2022)Zhang, Roller, Goyal, Artetxe, Chen, Chen, Dewan,
  Diab, Li, Lin, et~al.]{zhang2022opt}
Susan Zhang, Stephen Roller, Naman Goyal, Mikel Artetxe, Moya Chen, Shuohui
  Chen, Christopher Dewan, Mona Diab, Xian Li, Xi~Victoria Lin, et~al.
\newblock Opt: Open pre-trained transformer language models.
\newblock \emph{arXiv preprint arXiv:2205.01068}, 2022.

\bibitem[Zhou et~al.(2022)Zhou, Loy, and Dai]{zhou2022extract}
Chong Zhou, Chen~Change Loy, and Bo~Dai.
\newblock Extract free dense labels from clip.
\newblock In \emph{European Conference on Computer Vision}, pp.\  696--712.
  Springer, 2022.

\bibitem[Zhu et~al.(2023)Zhu, Chen, Tian, Tao, Su, Yang, Huang, Li, Lu, Wang,
  et~al.]{zhu2023ghost}
Xizhou Zhu, Yuntao Chen, Hao Tian, Chenxin Tao, Weijie Su, Chenyu Yang, Gao
  Huang, Bin Li, Lewei Lu, Xiaogang Wang, et~al.
\newblock Ghost in the minecraft: Generally capable agents for open-world
  enviroments via large language models with text-based knowledge and memory.
\newblock \emph{arXiv preprint arXiv:2305.17144}, 2023.

\end{thebibliography}
